\documentclass[10pt,twocolumn,letterpaper]{article}
\pdfoutput=1

\usepackage[titletoc,title]{appendix}
\usepackage{amsfonts}
\usepackage{amsmath}
\usepackage{amssymb}
\usepackage{amstext}
\usepackage{bbm}
\usepackage{booktabs}
\usepackage{chngcntr}
\usepackage{colortbl}
\usepackage{cvpr}
\usepackage{float}
\usepackage[T1]{fontenc}
\usepackage[utf8]{inputenc}
\usepackage{graphicx}
\usepackage{mathtools}
\usepackage[inline,shortlabels]{enumitem}
\usepackage{pgfplots}
\usepackage{subcaption}
\usepackage{tikz}
\usepackage{times}
\usepackage{xspace}

\newcommand\numberthis{\addtocounter{equation}{1}\tag{\theequation}}

\newcommand{\RANSAC}{\textsc{ransac}\xspace}

\newcommand{\LORANSAC}{\textsc{lo-ransac}\xspace}

\newcommand{\M}[1]{\mathtt{#1}}

\pgfplotsset{compat=newest}
\usetikzlibrary{plotmarks}
\usetikzlibrary{external}
\usepgfplotslibrary{patchplots}

\makeatletter
\newenvironment{customlegend}[1][]
{%
    \begingroup
    \pgfplots@init@cleared@structures
    \pgfplotsset{#1}%
}
{
  \pgfplots@createlegend
    \endgroup
}

\def\addlegendimage{\pgfplots@addlegendimage}
\makeatother

\newlength\fwidth 

\newcommand{\ma}[1]{\ensuremath{\mathtt{#1}}\xspace}
\newcommand{\ve}[1][x]{\ensuremath{\mathbf{#1}}\xspace}
\newcommand{\inv}{{-1}}
\newcommand{\T}{{\!\top}}

\newcommand{\vX}{\ensuremath{\ve[X]_{i}}\xspace}
\newcommand{\vXp}{\ensuremath{\ve[X]^{\prime}_{i}}\xspace}
\newcommand{\vx}{\ensuremath{\ve[x]_{i}}\xspace}
\newcommand{\vxp}{\ensuremath{\ve[x]^{\prime}_{i}}\xspace}
\newcommand{\vxd}{\ensuremath{\ve[\tilde{x}]_{i}}\xspace}
\newcommand{\vxdp}{\ensuremath{\ve[\tilde{x}]^{\prime}_{i}}\xspace}

\newcommand{\mH}{\ensuremath{\ma{H}}\xspace}
\newcommand{\mHinf}{\ensuremath{\ma{H}_{\infty}}\xspace}

\newcommand{\mHinfhat}{\ensuremath{\ma{\hat{H}_{\infty}}}\xspace}

\newcommand{\mP}{\ensuremath{\ma{P}}\xspace}

\newcommand{\vl}{\ensuremath{\ve[l]}\xspace}

\newcommand{\vu}{\ensuremath{\ve[u]}\xspace}
\newcommand{\vv}{\ensuremath{\ve[v]}\xspace}

\newcommand{\mA}{\ensuremath{\ma{A}}\xspace}

\newcount\colveccount
\newcommand*\colvec[1]{
        \global\colveccount#1
        \begin{pmatrix}
        \colvecnext
}
\def\colvecnext#1{
        #1
        \global\advance\colveccount-1
        \ifnum\colveccount>0
                \\
                \expandafter\colvecnext
        \else
                \end{pmatrix}
        \fi
}

\newtoks\rowvectoks
\newcommand{\rowvec}[2]{%
  \rowvectoks={#2,}\count255=#1\relax
  \advance\count255 by -1
  \rowvecnexta}
\newcommand{\rowvecnexta}{%
  \ifnum\count255>0
    \expandafter\rowvecnextb
  \else
    \setlength\arraycolsep{1pt}     
    \begin{pmatrix}\the\rowvectoks\end{pmatrix}
  \fi}
\newcommand\rowvecnextb[1]{%
  \ifnum\count255>1     
    \rowvectoks=\expandafter{\the\rowvectoks&#1,}%
  \else
    \rowvectoks=\expandafter{\the\rowvectoks&#1}%
  \fi
    \advance\count255 by -1
    \rowvecnexta}

\newcommand*\rot{\rotatebox{90}}

\definecolor{Gray}{gray}{0.9}
\newcolumntype{a}{>{\columncolor{Gray}}c}

\newcommand{\ra}[1]{\renewcommand{\arraystretch}{#1}}

\usepackage[pagebackref=true,breaklinks=true,letterpaper=true,colorlinks,bookmarks=false]{hyperref}

\cvprfinalcopy 

\def\cvprPaperID{3217} 

\def\mytitle{Radially-Distorted Conjugate Translations}

\ifcvprfinal\fi
\begin{document}

\title{\mytitle}
\author{James Pritts\textsuperscript{1}\quad\quad Zuzana Kukelova\textsuperscript{1}\quad\quad Viktor Larsson\textsuperscript{2}\quad\quad Ond{\v r}ej Chum\textsuperscript{1}\\
{\small Visual Recognition Group, CTU in Prague\textsuperscript{1} \quad\quad Centre for Mathematical Sciences, Lund University\textsuperscript{2}}
}

\maketitle

\begin{abstract}
This paper introduces the first minimal solvers that jointly solve for
affine-rectification and radial lens distortion from coplanar repeated
patterns.  Even with imagery from moderately distorted lenses, plane
rectification using the pinhole camera model is inaccurate or
invalid. The proposed solvers incorporate lens distortion into the
camera model and extend accurate rectification to wide-angle imagery,
which is now common from consumer cameras. The solvers are derived
from constraints induced by the conjugate translations of an imaged
scene plane, which are integrated with the division model for radial
lens distortion. The hidden-variable trick with ideal saturation is
used to reformulate the constraints so that the solvers generated by
the Gr{\"o}bner-basis method are stable, small and fast.

Rectification and lens distortion are recovered from either one
conjugately translated affine-covariant feature or two independently
translated similarity-covariant features. The proposed solvers are
used in a \RANSAC-based estimator, which gives accurate rectifications
after few iterations. The proposed solvers are evaluated against the
state-of-the-art and demonstrate significantly better rectifications
on noisy measurements.  Qualitative results on diverse imagery
demonstrate high-accuracy undistortions and rectifications. The source
code is publicly
available\footnote{\url{https://github.com/prittjam/repeats}}.

\end{abstract}

\section{Introduction}
Scene-plane rectification is used in many classic computer-vision
tasks, including single-view 3D reconstruction, camera calibration,
grouping coplanar symmetries, and image editing
\cite{Wu-CVPR11,Pritts-CVPR14,Lukac-ACMTG17}. In particular, the
affine rectification of a scene plane transforms the camera's
principal plane so that it is parallel to the scene plane. This
restores the affine invariants of the imaged scene plane, which
include parallelism of lines and translational symmetries
\cite{Hartley-BOOK04,Pritts-CVPR14}. There is only an affine
transformation between the affine-rectified imaged scene plane and its
real-world counterpart. The removal of the effects of perspective
imaging is helpful to understanding the geometry of the scene plane.

Wide-angle imagery that has significant lens distortion is common
since consumer photography is now dominated by mobile-phone and
GoPro-type cameras. High-accuracy rectification from wide-angle
imagery is not possible with only pinhole camera models
\cite{Kukelova-CVPR15,Wildenauer-BMVC13}. Lens distortion can be
estimated by performing a camera calibration apriori, but a fully
automated method is desirable.


\begin{figure}
\includegraphics[width=0.495\columnwidth]{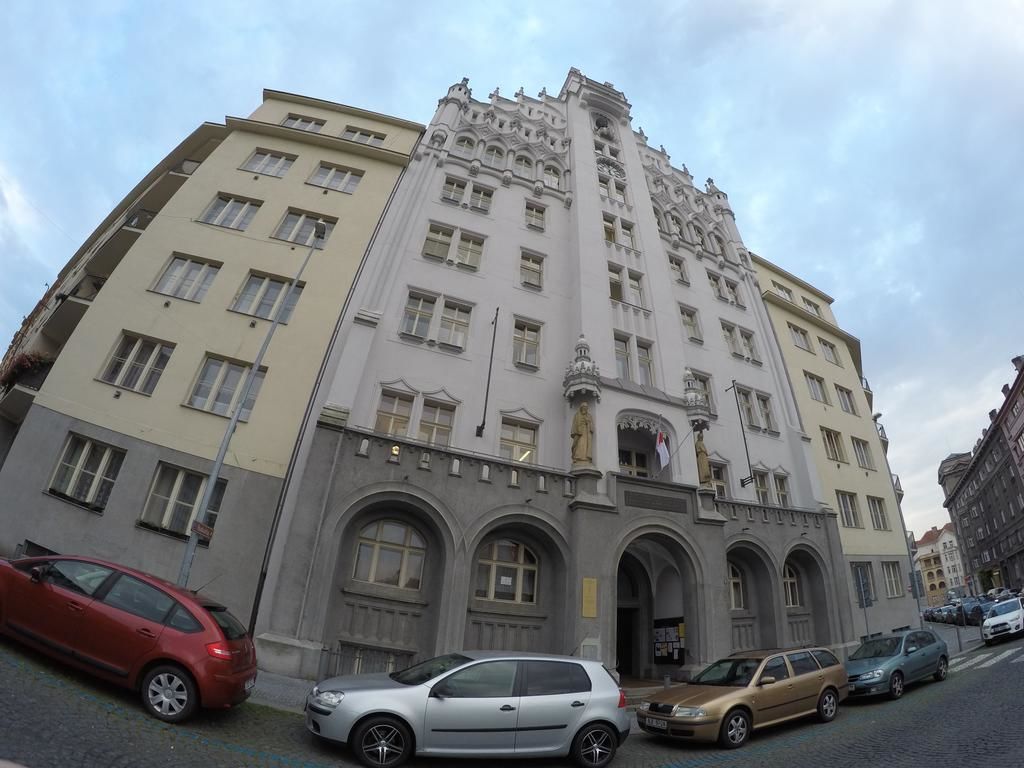}
\includegraphics[width=0.495\columnwidth]{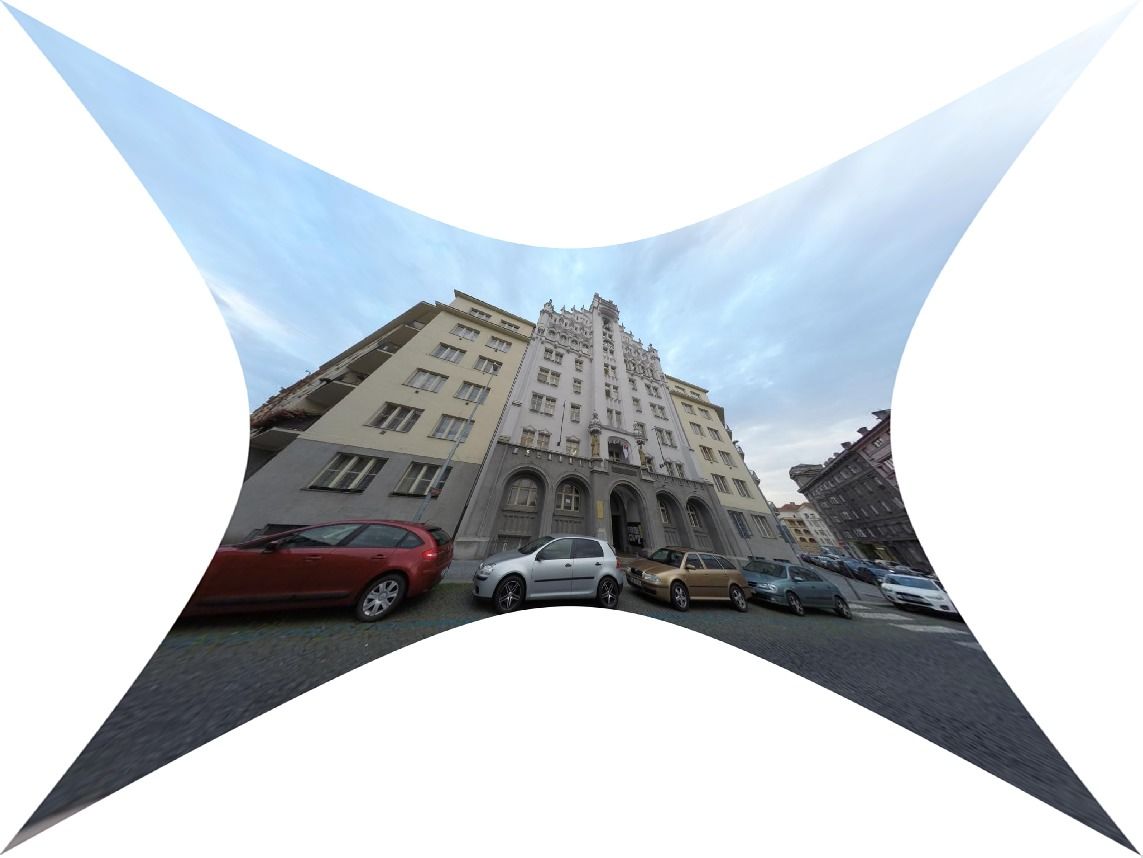}

\vspace{0.3cm}

\includegraphics[width=0.995\columnwidth]{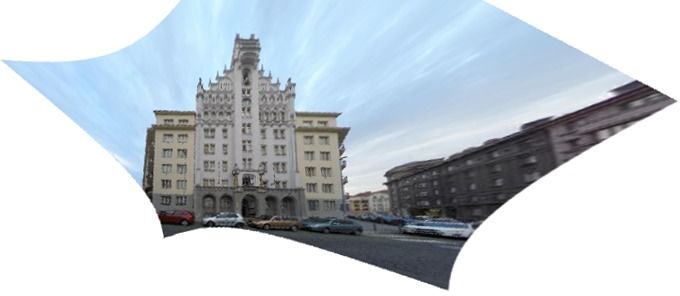}
\caption{Input (top left) is a distorted view of a scene plane, and the outputs (top right, bottom) are the the undistorted and rectified scene plane. The method is fully automatic.}
\label{fig:first}
\end{figure}

Several state-of-the-art planar-rectification methods assume a pinhole
camera model, which ignores the effect of lens distortion
\cite{Aiger-EG12,Chum-ACCV10,Lukac-ACMTG17,Zhang-IJCV12}. Pritts \etal
\cite{Pritts-CVPR14} attempt to upgrade the pinhole camera model with
radial lens distortion by giving an initial guess of the scene plane's
rectification that is consistent with a pinhole camera to a non-linear
optimization that incorporates a lens-distortion model.  However, even
with relaxed thresholds, a robust estimator (\ie \RANSAC) will discard
measurements that capture the most extreme effects of lens distortion,
especially around the boundary of the image, since these measurements
are not consistent with the pinhole-camera assumption. Thus, failing
to account for lens distortion while labeling the measurements as
outliers, as done during a \RANSAC iteration, can give biased fits
that underestimate the camera's lens-distortion
\cite{Kukelova-CVPR15}, which, in turn, reduces rectification accuracy.

\begin{figure}[t!]
\begin{minipage}{0.5\textwidth}
\end{minipage}
\begin{minipage}{0.10\textwidth}
\centering
narrow
\includegraphics[width=\textwidth]{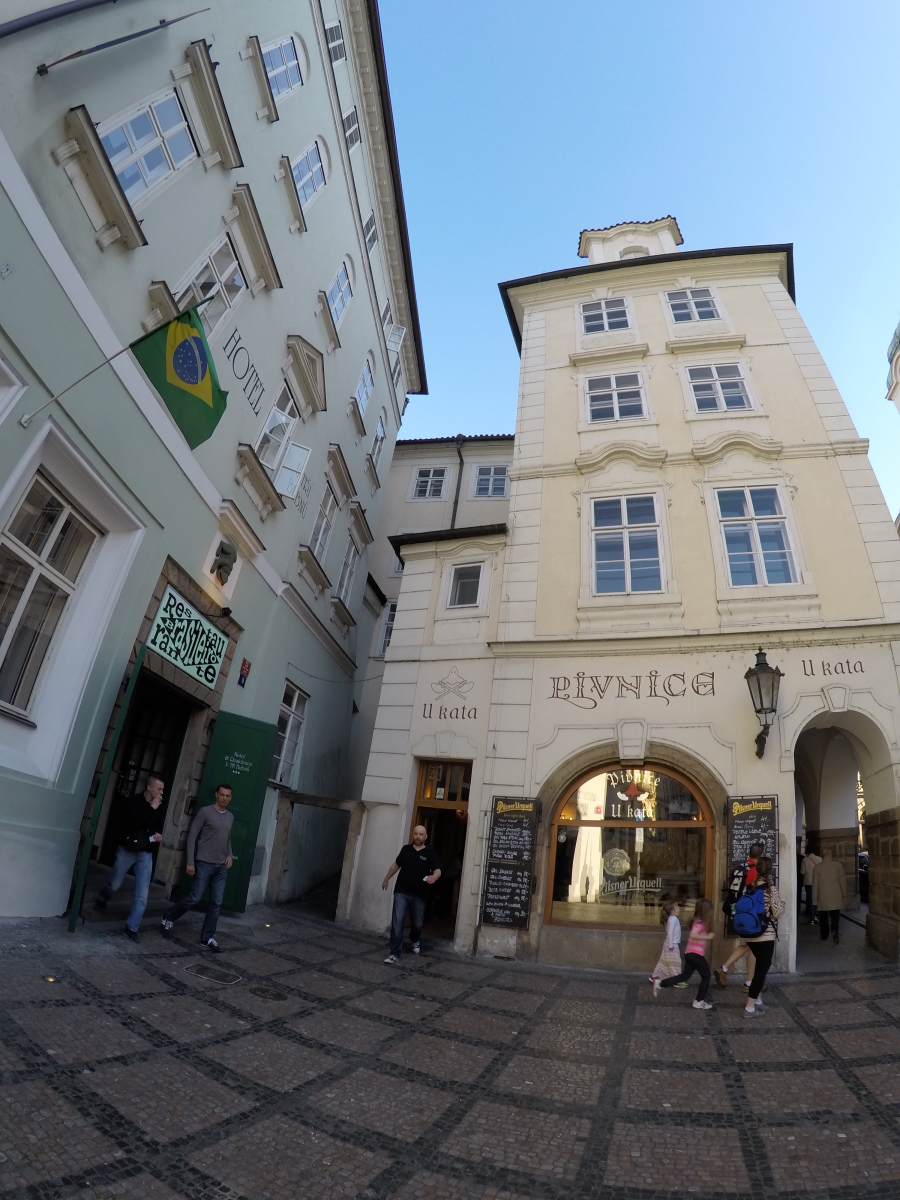}
\end{minipage}
\begin{minipage}{0.18\textwidth}
\centering
medium
\includegraphics[width=\textwidth]{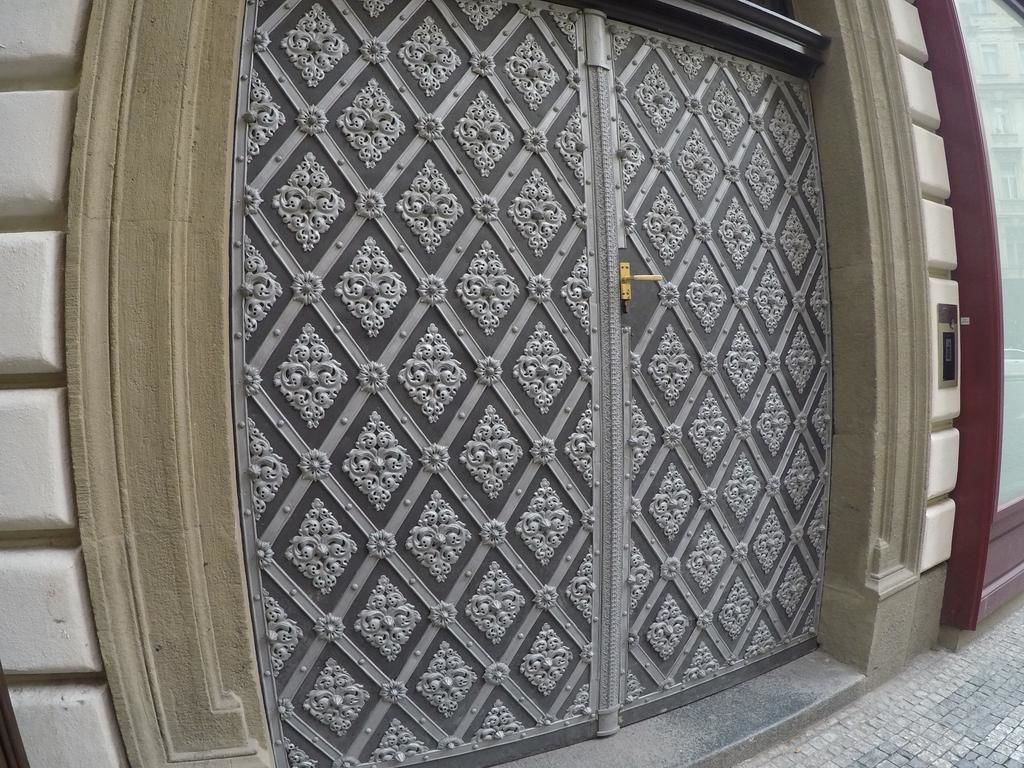}
\end{minipage}
\begin{minipage}{0.18\textwidth}
\centering
wide
\includegraphics[width=\textwidth]{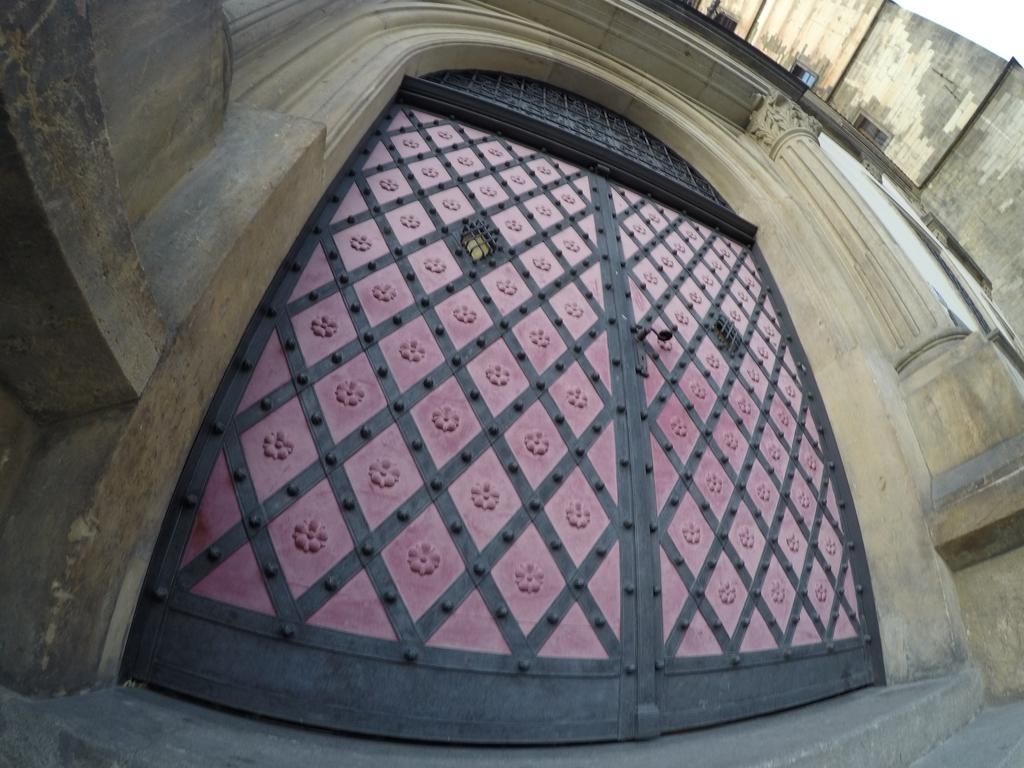}
\end{minipage}

\begin{minipage}{0.10\textwidth}
\centering
\includegraphics[width=\textwidth]{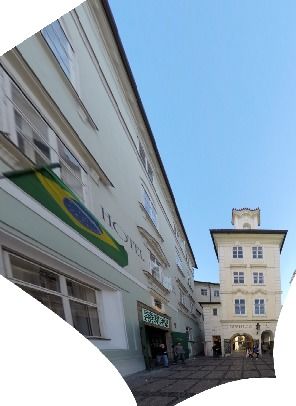}
\end{minipage}
\begin{minipage}{0.18\textwidth}
\centering
\includegraphics[width=\textwidth]{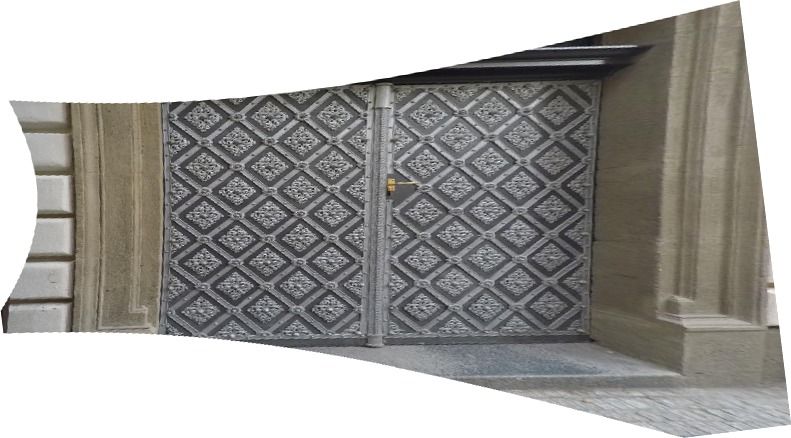}
\end{minipage}
\begin{minipage}{0.18\textwidth}
\centering
\includegraphics[width=\textwidth]{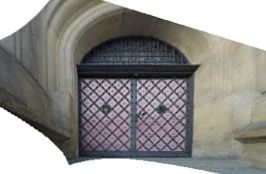}
\end{minipage}
\caption{\emph{GoPro Hero 4 imagery.}  (top row) Input images taken at different field-of-view settings. (bottom row) Rectified results.}
\label{fig:gopro}
\end{figure}

This paper introduces the first minimal solvers that jointly solve for
the affine rectification of an imaged scene plane and a camera's
radial lens distortion. The solvers are derived from constraints
induced by the conjugate translations of an imaged scene plane (see
Sec.~\ref{sec:problem_formulation} for details), which are integrated
with the division model of radial distortion \cite{Fitzgibbon-CVPR01}.
Despite the simple formulation of the division model, it is accurate
for even wide-angle lenses~\cite{Fitzgibbon-CVPR01}.  In addition, the
solvers estimate the vanishing translation direction of the
corresponded points used for input.

Two types of solvers are introduced: \emph{one-direction solvers},
which require 3 coplanar point correspondences that translate in the
same direction, and \emph{two-direction solvers}, which require 4
coplanar point correspondences, 2 of which translate in one-direction
and the remaining 2 in a different direction. Covariant feature
detectors are used to extract the needed point
correspondences \cite{Lowe-IJCV04,Matas-BMVC02,Mikolajczyk-IJCV04,Mishkin-ARXIV17,Vedaldi-SOFTWARE08}. The
solvers are used in a \RANSAC-based framework for robust rectification
estimation. With one or two-correspondence sampling, an accurate
undistortion and rectification is quickly recovered, even for
difficult scenes.

Fitzgibbon used a one-parameter division model to develop a minimal
solver for jointly estimating lens distortion with a fundamental
matrix or homography \cite{Fitzgibbon-CVPR01}.
Kukelova \etal \cite{Kukelova-CVPR15} proposed an extension
to \cite{Fitzgibbon-CVPR01} for homographies to model two-views from
cameras with different radial lens distortions. These two-view solvers
can jointly estimate lens distortion and conjugate translations, but
are overparameterized for this task, which can result in inaccurate
estimates as is shown by the synthetic experiments in
Sec.~\ref{sec:synthetic_experiments}. Wildenauer \etal \cite{Wildenauer-BMVC13}
and Antunes \etal
\cite{Antunes-CVPR17} are two methods that use constraints induced by imaged parallel
lines to jointly solve for their vanishing point and the division
model parameter, but but both require a multi-model estimation to
recover scene-plane rectification (\ie 2 consistent vanishing points).


The systems of polynomial equations induced from the constraints
arising from joint estimation of conjugate translation with the
division-model parameter are solved using an algebraic method based on
Gr{\"o}bner bases.  Automated solver-generators using the Gr{\"o}bner
basis method~\cite{Kukelova-ECCV08,Larsson-CVPR17} were recently used
to generate solvers for several problems in multi-view
geometry \cite{Kukelova-ECCV08,Larsson-ICCV17,Larsson-CVPR17,Kukelova-CVPR15}. However,
straightforward application of an automated solver-generator to the
proposed problem resulted in unstable solvers (see
Sec.~\ref{sec:synthetic_experiments}).  Therefore, we transformed the
constraints to simplify the structure of the systems of polynomial
equations, while explicitly accounting for the parasitic solutions
that arose from the new formulation. The new formulation resulted in
solvers with increased stability and speed.
 
The problem of rectification is closely coupled with the detection of
coplanar repeats in a classic chicken-and-egg scenario: rectification
is easy if the repeats are grouped, and repeats are more easily
grouped if the affine invariants of the rectified plane are available
\cite{Pritts-CVPR14}. Most methods tentatively group repeats from
their local texture, which is verified later by a hypothesized
rectification. Methods using this approach include Schaffalitzky \etal
\cite{Schaffalitzky-BMVC98}, which, similar to the solvers proposed in
this paper, uses constraints induced by conjugate translations to
recover the scene-plane's vanishing line, and Chum \etal
\cite{Chum-ACCV10}, which uses the constraint that coplanar repeats
are equiareal in the scene-plane's affine-rectified image. None of
these methods account for lens distortion, and do not perform well on
imagery with significant lens distortion (see
Sec.~\ref{sec:real_images}).

%

\section{Problem Formulation}
\label{sec:problem_formulation}
Assume that the scene plane $\pi$ and a camera's image plane $\pi'$
are related point-wise by the homography $\mP$, so that $\alpha_i
\vxp=\mP\vX$, where $\alpha_i$ is a scalar, $\vX \in \pi$ and $\vxp
\in \pi'$. Let $\vX$ and $\vXp$ be two points on the scene plane $\pi$
such that $\vXp-\vX=\ve[t]$. By encoding $\ve[t]$ in the homogeneous
translation matrix $\ma{T}$, the points $\vX$ and $\vXp$ as imaged by
camera $\mP$ can be expressed as
\begin{equation}
  \label{eq:conjugate_translation} \alpha_i\vxp=\mP\vXp=\mP\ma{T}\vX=\mP\ma{T}\mP^{\inv}\vx=\ma{H_{\ve[u]}}\vx,
\end{equation}
where the homography $\ma{H}_{\ve[u]}=\mP\ma{T}\mP^{\inv}$ is called a
conjugate translation because of the form of its matrix decomposition
and points $\vx$ and $\vxp$ are in correspondence with respect to the
conjugate translation $\ma{H}_{\ve[u]}$, which we denote
$\vx \leftrightarrow \vxp$ \cite{Hartley-BOOK04,Schaffalitzky-BMVC98}. Decomposing
$\ma{H}_{\ve[u]}$ into its projective components gives
\begin{align*}
  \alpha_i \vxp &= \mH_{\vu}\vx
  = \left[ \mP\ma{I}_3\mP^{\inv}+\mP\colvec{3}{t_x}{t_y}{1}\left[\mP^{-\T}\colvec{3}{0}{0}{1}\right]^{\T} \right] \vx \\
  &=
  [\ma{I}_3+s^{\vu}_i\ve[u]\ve[l]^{\T}] \cdot \vx \numberthis \label{eq:decomposition}
\end{align*}
where $\ma{I}_3$ is the $3\times 3$ identity matrix, and
\begin{itemize} 
\itemsep0em 
\item line \vl is the imaged scene plane's vanishing line,
\item point $\vu$ is the vanishing direction of translation, which must meet the vanishing line \vl, \ie, $\vl^{\T}\vu=0$,
\item and scalar $s^{\vu}_i$ is the magnitude of translation in the direction $\ve[u]$ for the point correspondence $\vxd \leftrightarrow \vxdp$ \cite{Schaffalitzky-BMVC98}.
\end{itemize}

Note that \eqref{eq:decomposition} holds only for points projected by
a pinhole camera viewing a scene plane, which is parameterized by the
homography $\mP$ as defined above. For every real camera, some amount
of radial distortion is always present, so for
\eqref{eq:decomposition} to hold, the measured image points $\vxd$ and
$\vxdp$ must first be undistorted. We use the one-parameter division
model to parameterize the radial lens distortion
\cite{Fitzgibbon-CVPR01}, which has the form
\begin{equation}
\label{eq:division_model}
f(\ve[\vxd],\lambda)=\rowvec{3}{\tilde{x}_i}{\tilde{y}_i}{1+\lambda(\tilde{x}_i^2+\tilde{y}_i^2)}^{\T},
\end{equation}
where the distortion center is given; \ie, $\tilde{x}_i,\tilde{y}_i$
are the center-subtracted measurements from a feature detector.

In this work we incorporate constraints induced by a conjugate
translation as derived in \eqref{eq:decomposition} with the division
model defined in \eqref{eq:division_model} to accurately rectify
imaged scene planes from lens-distorted cameras. Since the unknown
division model parameter is exclusively encoded in the homogeneous
coordinate, the relation for conjugate translations can be directly
augmented to model lens distortion, namely,
\begin{equation}
  \label{eq:distorted_conjugate_translation} \alpha_i f(\vxdp,\lambda)
  = \mH_{\ve[u]} f(\vxd,\lambda) =
  [\ma{I}_3+s^{\vu}_i\vu\ve[l]^{\T}] \cdot f(\vxd,\lambda),
\end{equation}
where $\alpha_i$ is some non-zero scalar, and
$\vxd \leftrightarrow \vxdp$ is a point correspondence.

\section{Solvers}
The model for radially-distorted conjugate translations
in \eqref{eq:distorted_conjugate_translation} defines the unknown
geometric quantities:
\begin{enumerate*}[(i)]\item division-model
parameter $\lambda$, \item scene-plane vanishing line
$\ve[l]=\rowvec{3}{l_1}{l_2}{l_3}^{\T}$,
\item vanishing translation direction
$\ve[u]=\rowvec{3}{u_1}{u_2}{u_3}^{\T}$ (see
Sec.~\ref{sec:two_direction} for the two-direction extensions),
\item scale of translation $s^{\vu}_i$ for correspondence $\vxd \leftrightarrow \vxdp$, 
\item and the scalar parameter $\alpha_i$.\end{enumerate*}

The solution for the vanishing line $\ve[l]$ is constrained to the
affine subspace $l_3=1$ of the real-projective plane, which makes it
unique.  This inhomogeneous choice of $\ve[l]$ is unable to represent
the pencil of lines that pass through the origin. If this degeneracy
is encountered, then the scale of $\ve[l]$ is fixed by setting $l_2=1$
instead. Solver variants for both constraints are generated for all of
the proposed solvers. In practice, this degeneracy is rarely
encountered. If the $l_3=1$ solver variant suffers from bad numerical
conditioning, then the $l_2=1$ variant can be activated and its
solutions tested for consensus with the measurements (see
Sec.~\ref{sec:ransac}). Without loss of generality the derivations
below assume that $l_3=1$.

The vanishing direction \vu must meet the vanishing line \vl, which
defines a subspace of solutions for \vu. The magnitude of
\vu is set to the translation scale $s^{\vu}_1$ of the first correspondence, which defines a  unique solution
\begin{equation}
\label{eq:orthogonality_constraint}
\vl^{\T}\vu=l_1u_1+l_2u_2+u_3=0 \quad \wedge \quad \|\vu\|=s^{\vu}_1.
\end{equation}
The relative scale of translation $\bar{s}^{\vu}_i$ for each
correspondence $\vxd \leftrightarrow \vxdp$ with respect to the
magnitude of $\|\vu\|$ is defined so that
$\bar{s}^{\vu}_{i}={s^{\vu}_{i}}/{\|\vu\|}$. Note that
$\bar{s}^{\vu}_{1}=1$.

In this paper we propose four different minimal solvers for different
variants of the problem of radially-distorted conjugate translations
based on different translation directions and relative scales
$\bar{s}^{\vu}_{i}$.  These variants are motivated by the types of
covariant feature detectors used to extract point
correspondences \cite{Lowe-IJCV04,Matas-BMVC02,Mikolajczyk-IJCV04,Mishkin-ARXIV17,Vedaldi-SOFTWARE08}.
Each affine-covariant feature defines an affine frame, \ie an ordered
triplet of points.  Thus, 1 affine-frame correspondence provides the 3
point correspondences that translate in the same direction with the
same scale. This is sufficient input for the one-directional
solvers. A visualization of the features is provided in
Fig.~\ref{fig:problem_difficulty} of the supplemental material.
In the case of similarity-covariant features, such as
DoG~\cite{Lowe-IJCV04}, only a similarity frame can be constructed. A
correspondence of similarity frames gives 2 point correspondences that
translate jointly. Two correspondences of similarity-covariant
features of different direction of the translation provide sufficient
constraints for the two directional solvers.

%





Two \emph{one-direction solvers} are proposed, which require 3 (2.5)
coplanar point correspondences that translate in the same direction.
The ``3-point'' solver $\mH3\vl\vu s_{\vu}\lambda$ assumes that two of
the point correspondences have the same scale of translation (i.e.\
$\bar{s}^{\vu}_{1} = \bar{s}^{\vu}_{2} = 1$), and the third point
correspondence has an unknown relative scale of the translation
$\bar{s}^{\vu}_{3}$.
The ``2.5-point'' solver $\mH2.5\vl\vu\lambda$ assumes that all 3
point correspondences have the same relative scales of translation,
i.e.\ $\bar{s}^{\vu}_{1} = \bar{s}^{\vu}_{2} = \bar{s}^{\vu}_{3} = 1$.

In addition, two \emph{two-direction solvers} are proposed, which
require 4 (3.5) coplanar point correspondences, 2 of which translate
in one-direction $\ve[u]$ and the remaining 2 in a different direction
$\ve[v]$.  Here the ``4-point'' solver $\mH4\vl\vu\vv s_{\vv}\lambda$
assumes that the first two point correspondences translate in the
direction $\ve[u]$ with the same relative scale of translation, i.e.,
$\bar{s}^{\vu}_{1} = \bar{s}^{\vu}_{2} = 1$.  The remaining two point
correspondence translate in the direction $\ve[v]$ with arbitrary
translation magnitudes, i.e., the relative scales of translations of
these two correspondences with respect to $\|\vv\|=s^{\vv}_3$ are
$\bar{s}^{\vv}_{3} = 1$ and an unknown relative scale
$\bar{s}^{\vu}_{4}$.

The ``3.5-point'' $\mH3.5\vl\vu\vv\lambda$ solver assumes that the
relative scales $\bar{s}^{\vu}_{1} = \bar{s}^{\vu}_{2} = 1$ with
respect to $\|\vu\|=s^{\vu}_1$ and $\bar{s}^{\vv}_{3}
= \bar{s}^{\vv}_{4} = 1$ with respect to $\|\vv\|=s^{\vv}_3$.

In all proposed solvers the scalar values $\alpha_i$ are eliminated
from
\eqref{eq:distorted_conjugate_translation}. This is done by
multiplying \eqref{eq:distorted_conjugate_translation} by the
skew-symmetric matrix $[f(\vxdp,\lambda)]_{\times}$. The fact that the
join of a point $\vx$ with itself $[\vx]_{\times}\vx$ is $\ve[0]$
gives,
\begin{align*}
&
\begin{bmatrix}
0 & -\tilde{w}^{\prime}_i & \tilde{y}^{\prime}_i \\
\tilde{w}^{\prime}_i & 0 & -\tilde{x}^{\prime}_i \\
-\tilde{y}^{\prime}_i & \tilde{x}^{\prime}_i & 0
\end{bmatrix} \\ & \times
\begin{bmatrix}
1+\bar{s}^{\vu}_{i}u_1l_1 & \bar{s}^{\vu}_{i}u_1l_2
& \bar{s}^{\vu}_{i}u_1 \\
\bar{s}^{\vu}_{i}u_2l_1   & 1+\bar{s}^{\vu}_{i}u_2l_2 & \bar{s}^{\vu}_{i}u_2  \\
\bar{s}^{\vu}_{i}u_3l_1   & \bar{s}^{\vu}_{i}u_3l_2   & 1+\bar{s}^{\vu}_{i}u_3
\end{bmatrix}  
\colvec{3}{\tilde{x}_i}{\tilde{y}_i}{\tilde{w}_i}
= \ve[0], \numberthis \label{eq:scalar_elimination}
\end{align*}
where $\tilde{w}_i=1+\lambda(\tilde{x}^2_i+\tilde{y}^2_i)$ and
$\tilde{w}^{\prime}_i=1+\lambda(\tilde{x}^{\prime
2}_i+\tilde{y}^{\prime 2}_i)$. The matrix equation
in \eqref{eq:scalar_elimination} contains three polynomial equations
from which only two are linearly independent, since the skew-symmetric
matrix $[f(\vxdp,\lambda)]_{\times}$ is rank two.

To solve the systems of polynomial equations resulting from the
presented problems, we use the Gr{\"o}bner basis
method~\cite{Cox-BOOK05}.
To generate efficient solvers we used the automatic generator of
Gr{\"o}bner basis solvers proposed
in \cite{Kukelova-ECCV08,Larsson-CVPR17}. However, for our problems
the coefficients of the input equations are not fully independent.
This means that using the default settings for the automatic generator
\cite{Kukelova-ECCV08,Larsson-CVPR17} that initialize the coefficients
of equations by random values from $\mathbb{Z}_p$ does not lead to
correct solvers. To obtain working Gr{\"o}bner basis solvers, one has
to create correct problems instances with values from $\mathbb{Z}_p$
for the automatic generator initialization.

The straightforward application of the automatic
generator~\cite{Kukelova-ECCV08,Larsson-CVPR17} to the needed
constraints with correct coefficients from $\mathbb{Z}_p$ resulted in
large templates and unstable solvers, especially for the two-direction
problems.  The Gr{\"o}bner basis solvers generated for the original
constraints have template matrices with sizes $80 \times 84$,
$74 \times 76$, $348 \times 354$, and $730 \times 734$ for the
$\mH2.5\vl\vu\lambda$, $\mH3\vl\vu s_{\vu}\lambda$,
$\mH3.5\vl\vu\vv\lambda$ and $\mH4\vl\vu\vv s_{\vv}\lambda$ problems,
respectively. Therefore, we use the hidden-variable
trick~\cite{Cox-BOOK05} to eliminate the vanishing translation
directions together with ideal saturation~\cite{Larsson-ICCV17} to
eliminate parasitic solutions. The reformulated constraints are
simpler systems in only 3 or 4 unknowns, and the solvers generated by
the Gr{\"o}bner basis method are smaller and more stable. The reduced
eliminiation template sizes are also summarized in
Sec.~\ref{sec:wall_clock} of the supplemental material. Next, we
describe the solvers based on the hidden-variable trick in more
detail.

\begin{table}[t!] 
\centering
\ra{1}
\begin{tabular}{@{} r c a a a a c c c @{}}
\toprule
& \rot{$\mH2\vl\ve[u]$} & \rot{$\mH2.5\vl\vu\lambda$} & \rot{$\mH3\vl\vu s_{\vu}\lambda$} &
\rot{$\mH3.5\vl\vu\vv\lambda$} & \rot{$\mH4\vl\vu\vv s_{\vv}\lambda$}
&  \rot{$\mH4\vl\gamma$}  & \rot{$\mH5\lambda$} & \rot{$\mH5\lambda_1\lambda_2$} \\
\midrule
Reference & \cite{Schaffalitzky-BMVC98} & & & &  & \cite{Chum-ACCV10} & \cite{Fitzgibbon-CVPR01} & \cite{Kukelova-ECCV08} \\
Distortion & & \checkmark & \checkmark & \checkmark & \checkmark &
& \checkmark & \checkmark \\
\mHinf & \checkmark & \checkmark & \checkmark & \checkmark & \checkmark & \checkmark
&  & \\
\# points & 2 & 2.5 & 3 & 3.5 & 4 & 4 & 5 & 5 \\
\# solutions & 1 & 4 & 2 & 6 & 4 & 1 & 18 & 5 \\
\bottomrule
\end{tabular}
\caption{Proposed solvers (grey) vs. state-of-the-art.} 
\label{table:solver_properties}
\end{table}

\subsection{One-Direction Solvers}
For the ``3-point'' one-direction $\mH3\vl\vu s_{\vu}\lambda$ solver
we have $\bar{s}^{\vu}_{1} = \bar{s}^{\vu}_{2} = 1$.  Therefore the
constraints~\eqref{eq:scalar_elimination} result in two pairs of
linearly independent equations without the scale parameter
$\bar{s}^{\vu}_{i}$ for $i=1,2$, and two linearly independent
equations with an unknown relative scale $\bar{s}^{\vu}_{3}$ for the
third point correspondence, i.e., \ $i=3$.  Additionally, we have the orthogonality constraint in~\eqref{eq:orthogonality_constraint}.  All together we have
seven equations in seven unknowns
($l_1,l_2,u_1,u_2,u_3,\bar{s}^{\vu}_{3},\lambda$).

Note, that these equations are linear with respect to the vanishing
translation direction $\ve[u]$.  Therefore, we can rewrite the seven
equations as
\begin{equation}
\M{M}(l_1,l_2,\bar{s}^{\vu}_{3},\lambda) \colvec{4}{u_1}{u_2}{u_3}{1}
= \ve[0] \label{eq:H3_HV}
\end{equation}
where $\M{M}(l_1,l_2,\bar{s}^{\vu}_{3},\lambda)$ is a $7\times 4$
matrix which elements are polynomials in
$(l_1,l_2,\bar{s}^{\vu}_{3},\lambda)$.

Since $\M{M}(l_1,l_2,\bar{s}^{\vu}_{3},\lambda)$ has a null vector, it
must be
rank deficient.  Therefore, all the $4 \times 4$ sub-determinants of
$\M{M}(l_1,l_2,\bar{s}^{\vu}_{3},\lambda)$ must equal zero.  This
results in ${{7}\choose{4}} = 35$ polynomial equations which only
involve four unknowns.

Unfortunately, the formulation~\eqref{eq:H3_HV} introduces a
one-dimensional family of false solutions. These are not present in
the original system and corresponds to solutions where the first three
columns of $\M{M}$ become rank deficient. In this case there exist
null vectors to $\M{M}$ where the last element of the vector is zero,
i.e. not on the same form as in \eqref{eq:H3_HV}.

These false solutions can be removed by saturating any of the $3\times
3$ sub-determinants from the first three columns of $\M{M}$.
The matrix $\M{M}$ has the following form,
\begin{equation} \small \label{eq:Mstructure}
\M{M}(l_1,l_2,\bar{s}^{\vu}_{3},\lambda) =
\begin{bmatrix}
m_{11} & m_{12} & 0 & m_{14} \\
m_{21} & m_{22} & 0 & m_{24} \\
m_{31} & 0 & m_{33} & m_{34} \\
m_{41} & 0 & m_{43} & m_{44} \\
m_{51} & m_{52} & 0 & m_{54} \\	
m_{61} & 0 & m_{63} & m_{64} \\
l_1 & l_2 & 1 & 0
\end{bmatrix}
\end{equation}   
where $m_{ij}$ are polynomials in $l_1,l_2,\bar{s}^{\vu}_3$ and
$\lambda$. We choose to saturate the $3\times 3$ sub-determinant
corresponding to the first, second and last row since it reduces to
only the top-left $2\times 2$ sub-determinant,
i.e. $m_{11}m_{22}-m_{12}m_{21}$, which is only a quadratic polynomial
in the unknowns. The other $3\times 3$ determinants are more
complicated and leads to larger polynomial solvers. Using the
saturation technique from Larsson \etal \cite{Larsson-ICCV17} we were
able to create a polynomial solver for this saturated ideal. The size
of the elimination template is $24 \times 26$. Note that without using
the hidden-variable trick the elimination template was $74\times 76$.

For the $\mH2.5\vl\vu\lambda$ solver we can use the same hidden-variable 
trick. In this case $\bar{s}^{\vu}_{1} = \bar{s}^{\vu}_{2}
= \bar{s}^{\vu}_{3} = 1$ and therefore the matrix $\M{M}$
in \eqref{eq:H3_HV} contains only three unknowns $l_1,l_2$ and
$\lambda$.  The minimal number of point correspondences necessary to
solve this problem is $2.5$. Therefore, for this problem we can drop
one of the equations from~\eqref{eq:scalar_elimination}, e.g., for
$i=3$, and the matrix $\M{M}$ in~\eqref{eq:H3_HV} has size $6 \times
4$. In this case all $4 \times 4$ sub-determinants of
$\M{M}$
result in 15 equations in 3 unknowns.

Similar to the 3 point case, this introduces a one-dimensional family
of false solutions. The matrix $\M{M}$ has a similar structure as
in \eqref{eq:Mstructure} and again it is sufficient to saturate
top-left $2\times 2$ sub-determinant. For this formulation we were
able to create a solver with template size $14 \times 18$ (compared
with $80\times 84$ without using hidden-variable trick)

\subsection{Two-Direction Solvers}
\label{sec:two_direction}
In the case of the two-direction $\mH4\vl\vu\vv s_{\vv}\lambda$ solvers, the input
equations for two vanishing translation directions
$\ve[u]=\rowvec{3}{u_1}{u_2}{u_3}^{\T}$ and
$\ve[v]=\rowvec{3}{v_1}{v_2}{v_3}^{\T}$ can be separated into two sets
of equations, i.e., the equations containing $\ve[u]$ and the equations
containing $\ve[v]$. Note that in this case we have two equations of
the form~\eqref{eq:orthogonality_constraint}, i.e., the equation for
the direction $\ve[u]$ and the equation for the direction $\ve[v]$ and
we have an unknown relative scale $\bar{s}^{\vv}_{4}$.  Therefore, the
final system of 10 equations in 10 unknowns can be rewritten using two
matrix equations as
\begin{equation}
\M{M}_1(l_1,l_2,\lambda) \colvec{4}{u_1}{u_2}{u_3}{1}
= \ve[0], \;\; \M{M}_2(l_1,l_2,\bar{s}^{\vv}_{4},\lambda) \colvec{4}{v_1}{v_2}{v_3}{1}
= \ve[0] \label{eq:H4_HV}
\end{equation}
where $\M{M}_1$ and $\M{M}_2$ are $5 \times 4$ matrices where the
elements are polynomials in $(l_1,l_2,\lambda)$ and
$(l_1,l_2,\bar{s}^{\vv}_{4},\lambda)$ respectively.

Again all $4 \times 4$ sub-determinants of $\M{M}_1$ and $\M{M}_2$
must concurrently equal zero.  This results in $5+5 = 10$ polynomial
equations in four unknowns $(l_1,l_2,\bar{s}^{\vv}_{4},\lambda)$.
In this case, only 39 additional false solutions arise from the
hidden-variable trick. The matrices $\M{M}_1$ and $\M{M}_2$ have a
similar structure as in \eqref{eq:Mstructure} and again  it is sufficient to saturate the top-left $2\times 2$ sub-determinants to
remove the extra solutions. By saturating these determinants we were able
to create a solver with template size $76\times 80$ (previously
$730\times 734$).

Finally, for the ``3.5-point'' two-direction $\mH3.5\vl\vu\vv\lambda$
solver  $\bar{s}^{\vu}_{1} = \bar{s}^{\vu}_{2} = 1$ 
and  
$\bar{s}^{\vv}_{3} = \bar{s}^{\vv}_{4} = 1$
so we can drop one of the equations from the
constraint~\eqref{eq:scalar_elimination}, e.g., for $i=4$.  Therefore,
the matrix $\M{M}_2$ from~\eqref{eq:H4_HV} has size $4\times 4$ and it
contains only 3 unknowns $(l_1,l_2,\lambda)$.  In this case all
$4 \times 4$ sub-determinants of $\M{M}_1$ and $\M{M}_2$ result in
$5+1 = 6$ polynomial equations in three unknowns $(l_1,l_2,\lambda)$.

For this case we get 18 additional false solutions. Investigations in
Macaulay2 \cite{M2} revealed that for this particular formulation it
was sufficient to only saturate the top-left $2\times 2$
sub-determinant of $\M{M}_1$ and the top-left element of
$\M{M}_2$. Saturating these we were able to create a polynomial solver
with a template size of $54 \times 60$ (previously $348\times 354$).

\section{Robust Estimation}
\label{sec:ransac}
The solvers are used in a \LORANSAC-based robust-estimation framework
~\cite{Chum-ACCV04}. Affine-covariant features are extracted from the
image for input to the solvers. Affine-covariant features are highly
repeatable on the same imaged scene texture with respect to
significant changes of viewpoint and illumination
\cite{Mikolajczyk-PAMI04}. Their proven robustness in the multi-view
correspondence problem makes them good candidates for representing the
local geometry of repeated textures. In particular, for the real-image
experiments in Sec.~\ref{sec:real_images}, we use the Maximally-Stable
Extremal Region and Hesssian-Affine detectors
\cite{Matas-BMVC02,Mikolajczyk-IJCV04}. The detections are
parameterized as 3 distinct points, which define an affine coordinate
frame in the image space \cite{Obdrzalek-BMVC02}. These detections are
visualized in Fig.~\ref{fig:problem_difficulty} of the supplemental
material.

Affine frames are labeled as repeated texture based on the similarity
of their appearance, which is given by the RootSIFT embedding of the
image patch local to the affine
frame \cite{Arandjelovic-CVPR12,Lowe-IJCV04}.  The RootSIFT
descriptors are agglomeratively clustered, which establishes pair-wise
tentative correspondences among the connected components linked by the
clustering. Each appearance cluster has some proportion of its members
that correspond to affine frames that give the geometry of imaged
repeated scene content, which are the \emph{inliers} of that
appearance cluster. The remaining affine frames are the
\emph{outliers}.

\LORANSAC samples pairs of affine frames from the appearance cluster,
which are inputted to the proposed minimal solvers.  Each pair of
affine frames across all appearance clusters has an equi-probable
chance of being drawn. The consensus with the minimal sample is
measured by the number of pairs of affine frames within appearance
groups that are consistent with the hypothesized model, normalized by
the size of each respective group. A non-linear optimizer
following~\cite{Pritts-CVPR14} is used as the local optimization step
of the \LORANSAC estimator.

%
%

\begin{figure}[t!]
  \raggedright \setlength\fwidth{0.75\columnwidth} \input{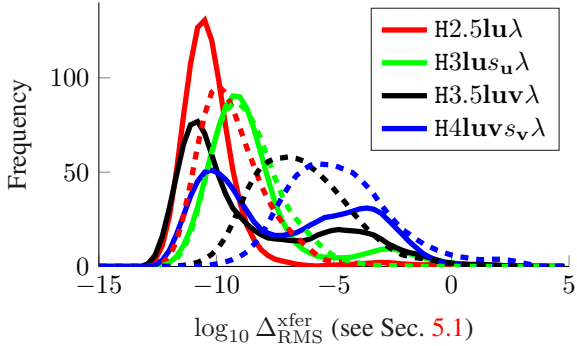}
\caption{\emph{Stability study.}  Hidden-variable
  trick solvers are solid; standard solvers are dashed. The
  $\log_{10}$ transfer error is reported. The hidden-variable trick
  increases stability.}  \label{fig:stability_study}
\end{figure}

\section{Synthetic Experiments}
\label{sec:synthetic_experiments}
The proposed solvers are evaluated across several benchmarks on
synthetic data against state-of-the-art solvers.
Included in the benchmarks are two single-view solvers:
$\mH2\ve[l]\ve[u]$ \cite{Schaffalitzky-BMVC98}, which also
incorporates constraints from conjugate translations, and
$\mH4\ve[l]\gamma$
\cite{Chum-ACCV10}, which solves for rectification and change-of-scale, and also two full-homography and radial distortion
solvers, $\mH5\lambda$ \cite{Fitzgibbon-CVPR01} and
$\mH5\lambda_1\lambda_2$ \cite{Kukelova-CVPR15}, which we use for
conjugate translation and lens-istortion estimation. The bench of
state-of-the-art solvers is summarized in
Table~\ref{table:solver_properties}).

The benchmarks are evaluated for 1000 synthetic images of 3D scenes
with known ground-truth parameters. A camera with a random but
realistic focal length is randomly oriented and positioned with
respect to a 10x10 square meter scene plane such that the plane is
mostly in the camera's field-of-view. Image resolution is set to
1000x1000 pixels. Conjugately translated affine frames are generated
on the scene plane such that their scale with respect to the scene
plane is realistic. This modeling choice reflects the use of
affine-covariant feature detectors for real images. The conjugately
translated features are distorted according to the division model,
and, for the sensitivity experiments, isotropic white noise is added
to the distorted affine frames at increasing levels. Performance is
characterized by the relative error of the estimated distortion
parameter and by the transfer and warp errors, which measure the
accuracies of the estimated conjugate translation and rectification
(see Sec.~\ref{sec:transfer_error}~-~\ref{sec:warp_error}). The
proposed solvers have an average solve time from 0.3 to 2 milliseconds
over the 1000 synthetic scenes (see also Sec.~\ref{sec:wall_clock} of
the supplemental material).

\subsection{Transfer Error}
\label{sec:transfer_error}
The geometric transfer error jointly measures the accuracy of an the
estimated conjugate translation and lens distortion. The scene plane
is tesselated by a 10x10 square grid of points $\{\,\vX\}$. Let the
translation on the scene plane induced by the noiseless pre-images of
the point correspondences used to estimate $\hat{\mH}_{\ve[u]}$ and
$\hat{\lambda}$ be $\ve[t]$. Then the grid points are translated by
$\ve[t]{}/\|\ve[t]\|$ to $\{\,\vX'\}$. The grid and its translation
are imaged by the ground-truth lens-distorted camera parameterized by
matrix $\mP$ and division-model parameter $\lambda$. The imaged grid
is given by $\vxd = f^{d}(\mP\vX,\lambda)$ and the translated grid by
$\vxdp = f^{d}(\mP\vX',\lambda)$, where $f^{d}$ is the the function
that transforms from pinhole points to radially-distorted points.
Then the geometric transfer error is defined as
\begin{equation}
  \label{eq:transfer_error} \Delta_i^{\mathrm{xfer}} =
 d(f^{d}([\ma{I}_3 + \frac{1}{\|\ve[t]\|}(\hat{\mH}_{\ve[u]}
- \ma{I}_3)]
f(\vxd,\hat{\lambda}_1),\hat{\lambda}_2),\vxdp),\end{equation} where
$d(\cdot,\cdot)$ is the Euclidean distance. All solvers except
$\mH5\lambda_1\lambda_2$ have the constraint that
$\hat{\lambda}_1=\hat{\lambda}_2$~\cite{Kukelova-CVPR15}. The
root-mean-square of transfer errors
$\Delta_{\mathrm{RMS}}^{\mathrm{xfer}}$ for correspondences
$\{\,(\vxd,\vxdp)\,\}$ is reported. For two-direction solvers, the
transfer error in the second direction is included in
$\Delta_{\mathrm{RMS}}^{\mathrm{xfer}}$. The transfer error is used in
the stability study, where the solvers are tested over varying
division model parameters and in the sensitivity study, where the
solvers are tested over varying noise levels with fixed division model
parameter. The solver $\mH4l\gamma$ of \cite{Chum-ACCV10} does not
estimate conjugate translations, so it is not reported.  For a
derivation of \eqref{eq:transfer_error} see
Sec.~\ref{sec:xfer_error_details} in the supplementary material.

\subsection{Numerical Stability}
\label{sec:stability}
The stability study measures the RMS transfer error of solvers (see
Sec.~\ref{sec:transfer_error}) for noiseless affine-frame
correspondences across realistic scene and camera configurations
generated as described in the introduction to this section. The
ground-truth parameter of the division model $\lambda$ is drawn
uniformly at random from the interval $[-6,0]$. For a reference, the
division parameter of $\lambda=-4$ is typical for wide field-of-view
cameras like the GoPro where the image is normalized by
${1}\over{width+height}$.  Fig.~\ref{fig:stability_study} reports the
histogram of $\log_{10}$ RMS transfer errors.  For all new solvers we
evaluate a solver generated from constraints derived with (solid
histogram) and without (dashed histogram) the hidden-variable
trick. The hidden-variable trick significantly improves the stability
of the proposed solvers.  The increased stabilities of the
hidden-variable solvers most likely result from the reduced size of
the G-J elimination problems needed by these solvers. The
hidden-variable solvers are used for the remainder of the experiments.

\begin{figure*}[t!]
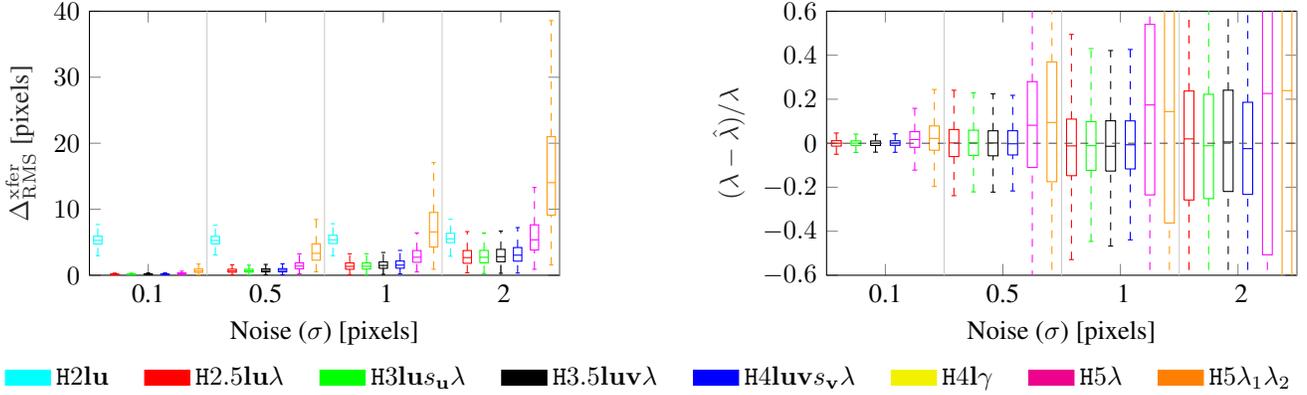

\raggedright
    \setlength\fwidth{0.75\columnwidth} \input{fig/rms_sensitivity.tikz} 
\hfill  
    \setlength\fwidth{0.75\columnwidth} \input{fig/lambda_sensitivity2.tikz}
  \centering \definecolor{mycolor1}{rgb}{0.00000,1.00000,1.0}
\definecolor{mycolor2}{rgb}{0.95,0.95,0.0}

\begin{tikzpicture}
\begin{customlegend}
[legend columns=-1,
legend style={draw=none,/tikz/every even column/.append style={column sep=0.4cm}},
legend entries={$\mH2\vl\vu$,$\mH2.5\vl\vu\lambda$,$\mH3\vl\vu s_{\vu}\lambda$,$\mH3.5\vl\vu\vv\lambda$,$\mH4\vl\vu\vv s_{\vv}\lambda$,$\mH4\vl\gamma$,$\mH5\lambda$,$\mH5\lambda_1\lambda_2$}]
    \addlegendimage{mycolor1,fill=mycolor1,area legend}            
    \addlegendimage{red,fill=red,area legend}            
    \addlegendimage{green,fill=green,area legend}            
    \addlegendimage{black,fill=black,area legend}                
    \addlegendimage{blue,fill=blue,area legend}          
    \addlegendimage{mycolor2,fill=mycolor2,area legend} 
    \addlegendimage{magenta,fill=magenta,area legend}            
    \addlegendimage{orange,fill=orange,area legend}            
\end{customlegend}
\end{tikzpicture} \caption{Comparison of the
  transfer error (left, see Sec.~\ref{sec:transfer_error}) and the
  relative radial distortion error (right) after 25 iterations of a
  simple \RANSAC for different solvers over increasingly noisy
  measurements for 1000 scenes.}
\label{fig:sensitivity_study}
\end{figure*}

\subsection{Noise Sensitivity}
The proposed and state-of-the-art solvers solvers are tested with
increasing levels of white noise added to the affine correspondences
induced by the ground-truth conjugate translation and lens distortion
parameter. The white noise is parameterized by the standard-deviation
of a zero-mean isotropic Gaussian distribution, and the solvers are
tested at noise levels of $\sigma \in \{0.1,0.5,1,2\}$. The ground
truth division model parameter is set to $\lambda=-4$, which is
typical for GoPro-type imagery. The solvers are wrapped
in \begin{figure}[H] \setlength\fwidth{0.75\columnwidth} \input{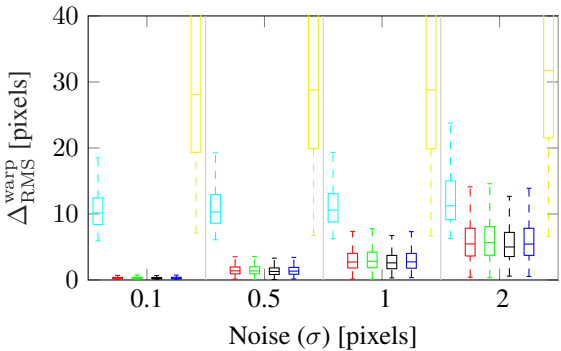} \caption{Warp-error
comparison (see Sec.~\ref{sec:warp_error}) after 25 iterations of a
simple \RANSAC for different solvers over increasingly noisy
measurements for 1000 scenes.}
\label{fig:warp}
\end{figure} \noindent a simple RANSAC-loop, which minimizes the RMS transfer error
$\Delta_{\mathrm{RMS}}^{\mathrm{xfer}}$ over 25 sampled affine-frame
correspondences. Results are calculated from the estimate given
by \RANSAC and summarized for the 1000 generated scenes as
boxplots. The interquartile range is contained within the extents of a
box, and the median is the horizontal line dividing the box.

As shown in Fig.~\ref{fig:sensitivity_study} the proposed solvers give
the most accurate joint estimation of conjugate translation and
division model parameter as measured by the RMS transfer error
$\Delta_{\mathrm{RMS}}^{\mathrm{xfer}}$ and relative error of
estimated divsion model parameter.  As expected, the minimal
parameterization of the radially-distorted conjugate translation
solvers---$\mH2.5\vl\vu\lambda,\mH3\vl\vu
s_{\vu}\lambda,\mH3.5\vl\vu\vv\lambda,\mH4\vl\vu\vv
s_{\vv}\lambda$---show significantly less sensitivity to noise than
the overparmeterized radially-distorted homography solvers
$\mH5\lambda$ and $\mH5\lambda_1\lambda_2$ for both measures. The
solver $\mH2\vl\ve[u]$ shows significant bias (see the transfer error
boxplots in Fig.~\ref{fig:sensitivity_study}) since it does not model
lens distortion.

\subsection{Warp Error}
\label{sec:warp_error}
Since the accuracy of scene-plane rectification is a primary concern,
we also report the warp error for rectifying homographies proposed by
Pritts \etal~\cite{Pritts-BMVC16}, which we augment with the division
model for radial lens distortion.  A rectifying homography \mHinf of
an imaged scene plane is constructed from its vanishing line
$\ve[l]$ (see \cite{Hartley-BOOK04}).
A round trip between the image space and rectified space is made by
undistorting and rectifying imaged coplanar points by the estimated
lens distortion $\hat{\lambda}$ and rectifying homography \mHinfhat
and then re-warping and distorting the rectified points into the image
by a synthetic camera constructed from the ground-truth lens
distortion $\lambda$ and rectifying homography $\mHinf$. Ideally, the
synthetic camera constructed from the truth would project the
undistorted and rectified points onto the original points.

Note that there is an affine ambiguity, denoted \mA, between \mHinfhat
and
\mHinf, which is folded into the expression for the synthetic
camera, namely $\mP(\mA)=(\mA\mHinf)^{\inv}$, and estimated during
computation of the warp error,
\begin{equation} 
  \label{prg:warp_residual} \Delta^{\mathrm{warp}}=\min_{\hat{\mA}} \sum_{i}
  d^2(\vxd,f^d(\mP(\hat{\mA})\mHinfhat
  f(\vxd,\hat{\lambda})),\hat{\lambda}),
\end{equation}
where $d(\cdot,\cdot)$ is the Euclidean distance, and $\{
\vxd \}$ are the imaged grid points of the scene-plane tesselation 
as defined in Sec.~\ref{sec:transfer_error}.  The root mean square
warp error for $\{\,\vxd \,\}$ is reported and denoted as
$\Delta^{\mathrm{warp}}_{\mathrm{RMS}}$.  The vanishing line is not
directly estimated by solvers $\mH5\lambda$
of \cite{Fitzgibbon-CVPR01} and $\mH5\lambda_1\lambda_2$
of \cite{Kukelova-CVPR15}, so they are not reported.

\begin{figure*}[t!]
\begin{minipage}{0.22\textwidth}
\centering
Original image
\includegraphics[width=\textwidth]{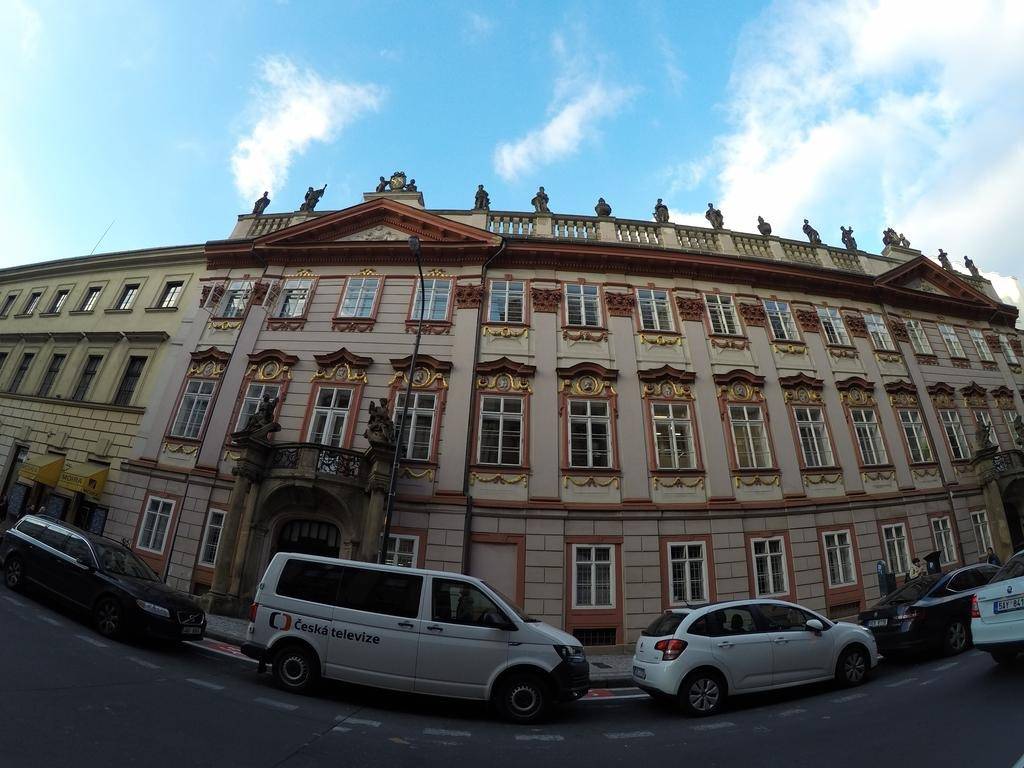}
\end{minipage}
\begin{minipage}{0.245\textwidth}
\centering
$\mH2\ve[l]\ve[u]$ + LO; $11.2 \%$ inliers
\includegraphics[width=\textwidth]{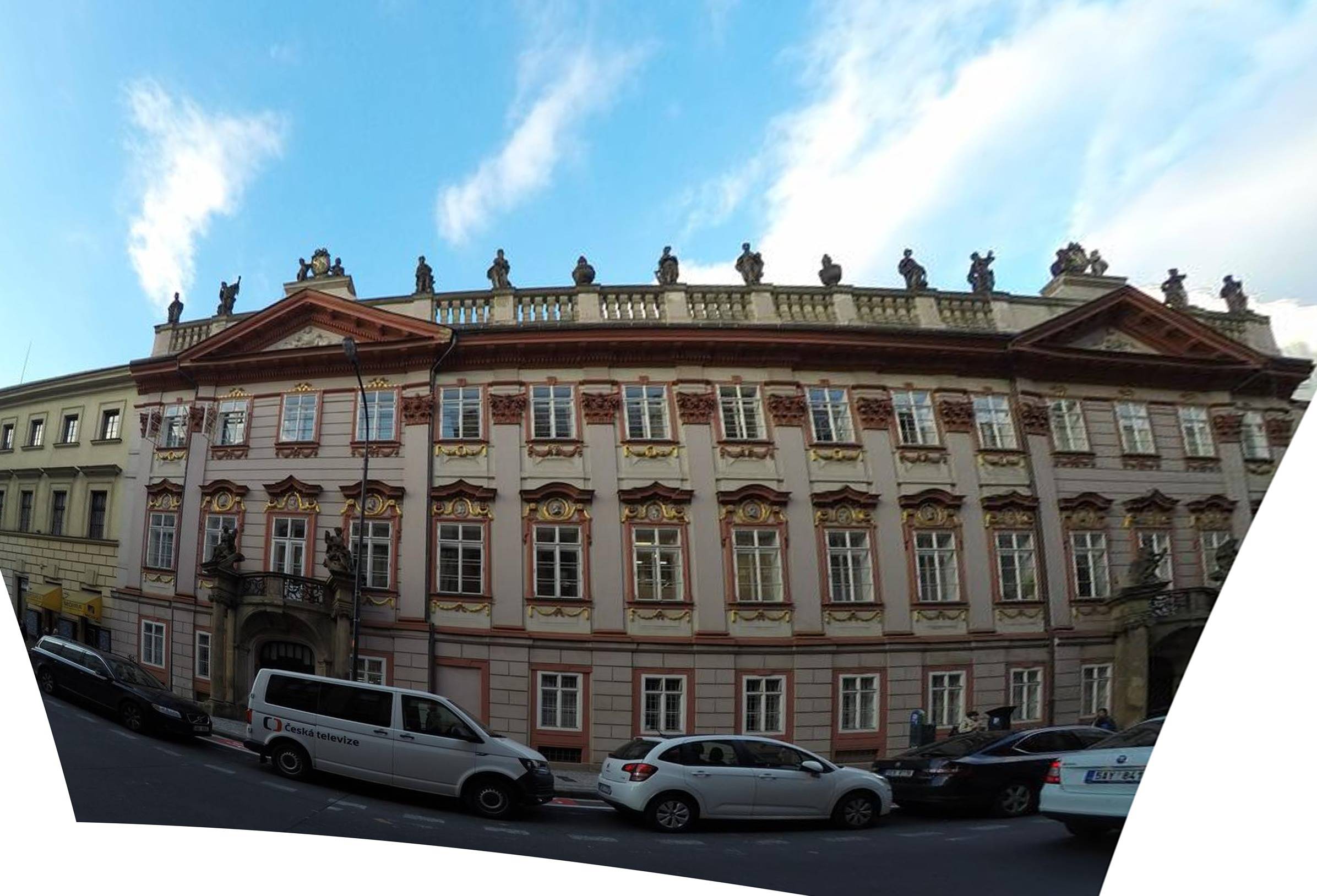}
\end{minipage}
\begin{minipage}{0.245\textwidth}
\centering
$\mH2.5\vl\vu\lambda$+LO; $20.4 \%$ inliers
\includegraphics[width=\textwidth]{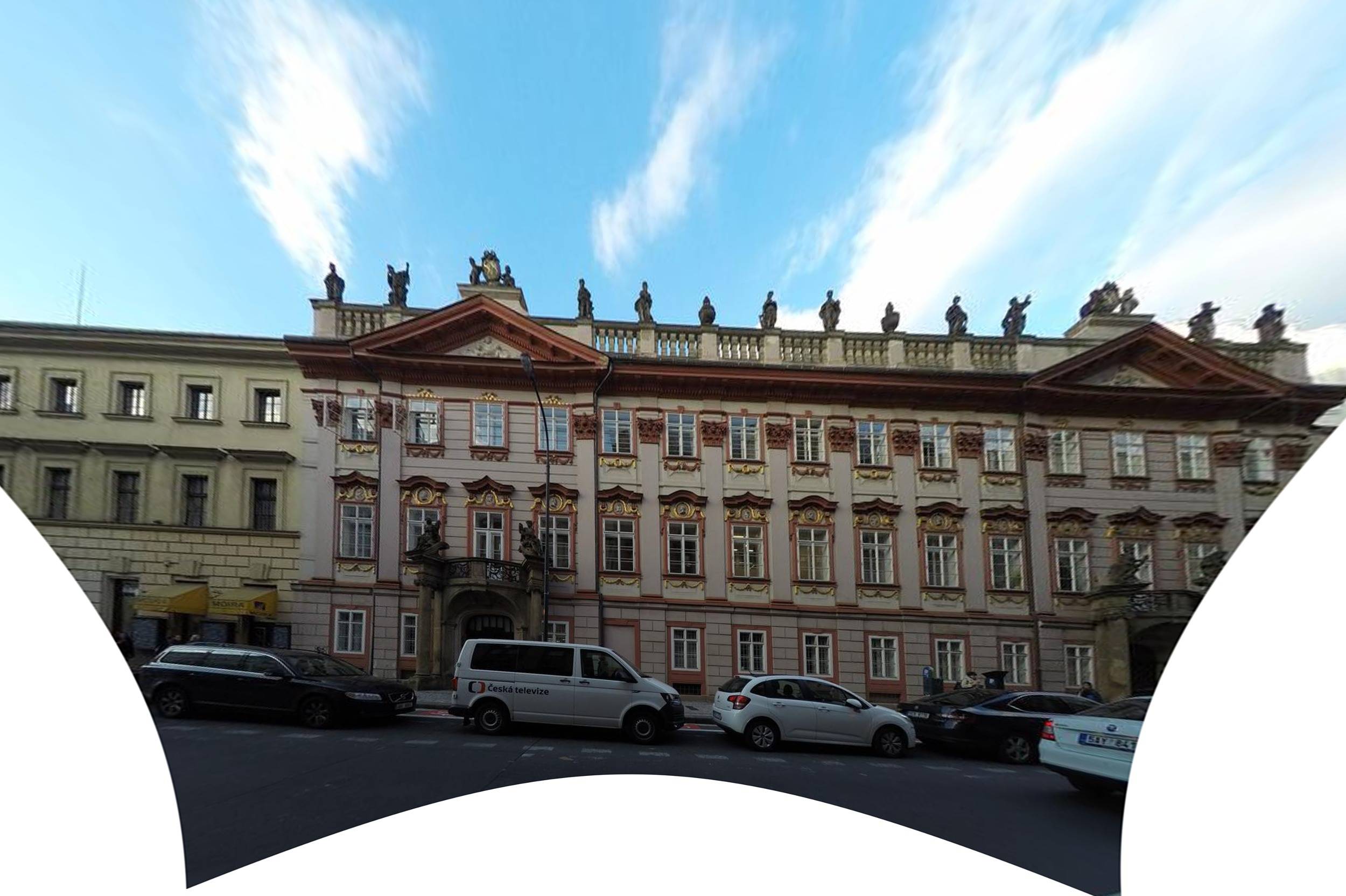}
\end{minipage}
\begin{minipage}{0.245\textwidth}
\centering
$\mH3.5\vl\vu\vv\lambda$+LO; $20.2 \%$ inliers
\includegraphics[width=\textwidth]{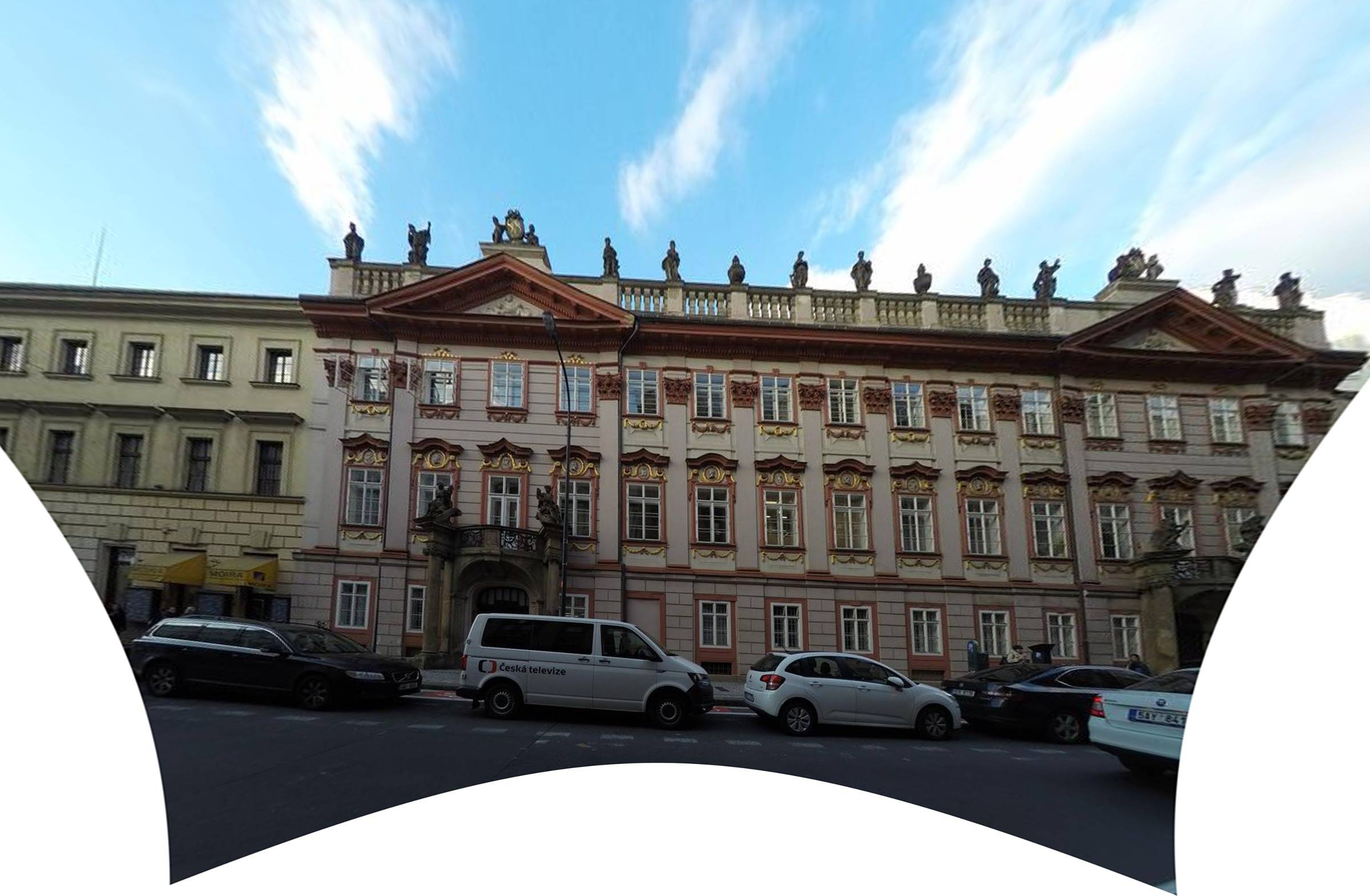}
\end{minipage}

\begin{minipage}{0.22\textwidth}
\centering
\includegraphics[width=\textwidth]{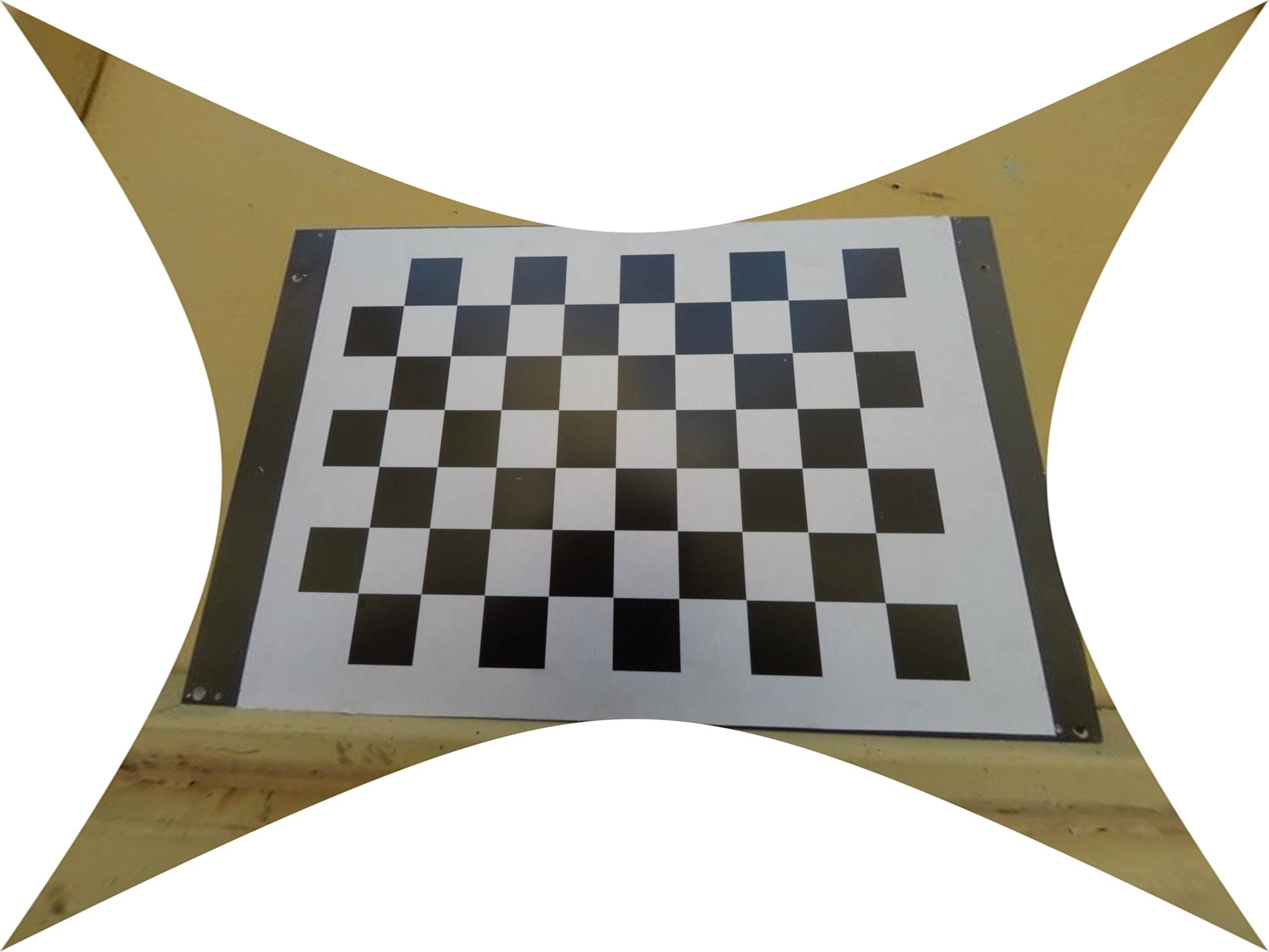}
\end{minipage}
\begin{minipage}{0.01\textwidth}
\hfill
\centering
\end{minipage}
\begin{minipage}{0.225\textwidth}
\centering
\includegraphics[width=\textwidth]{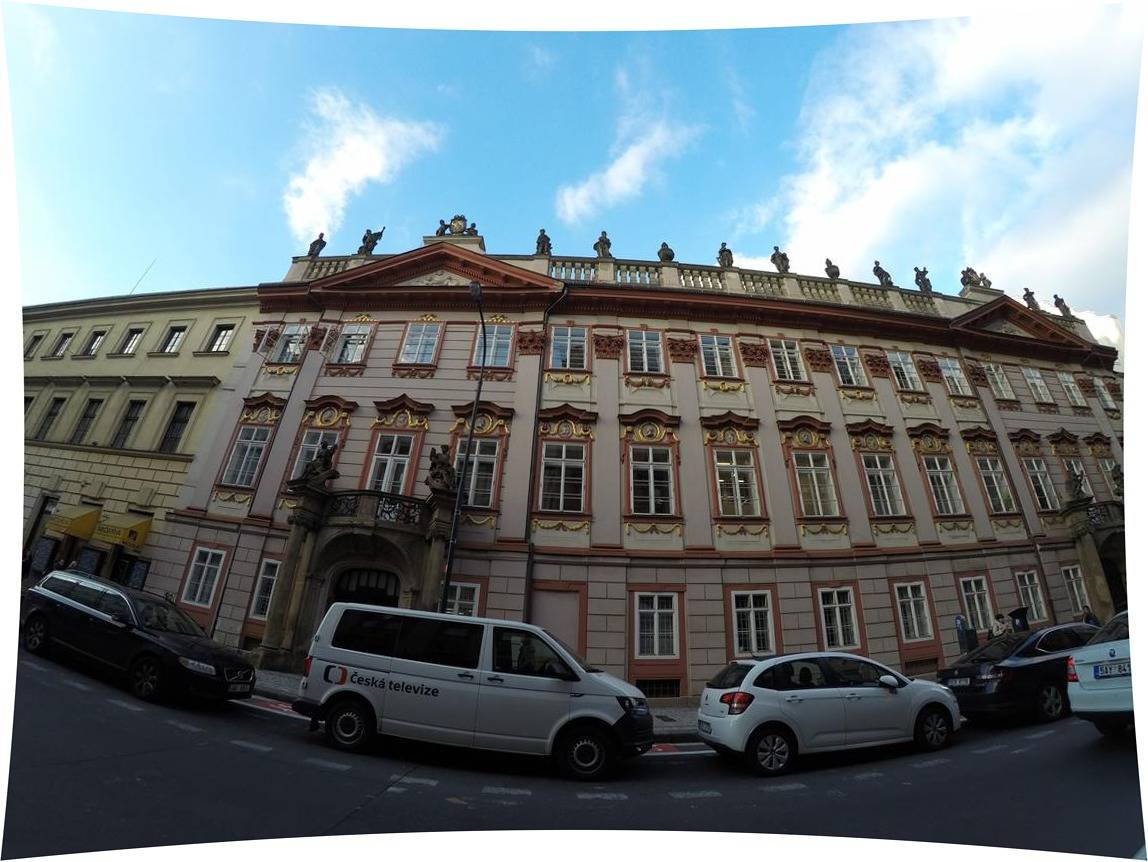}
\end{minipage}
\begin{minipage}{0.01\textwidth}
\hfill
\centering
\end{minipage}
\begin{minipage}{0.01\textwidth}
\hfill
\centering
\end{minipage}
\begin{minipage}{0.225\textwidth}
\centering
\includegraphics[width=\textwidth]{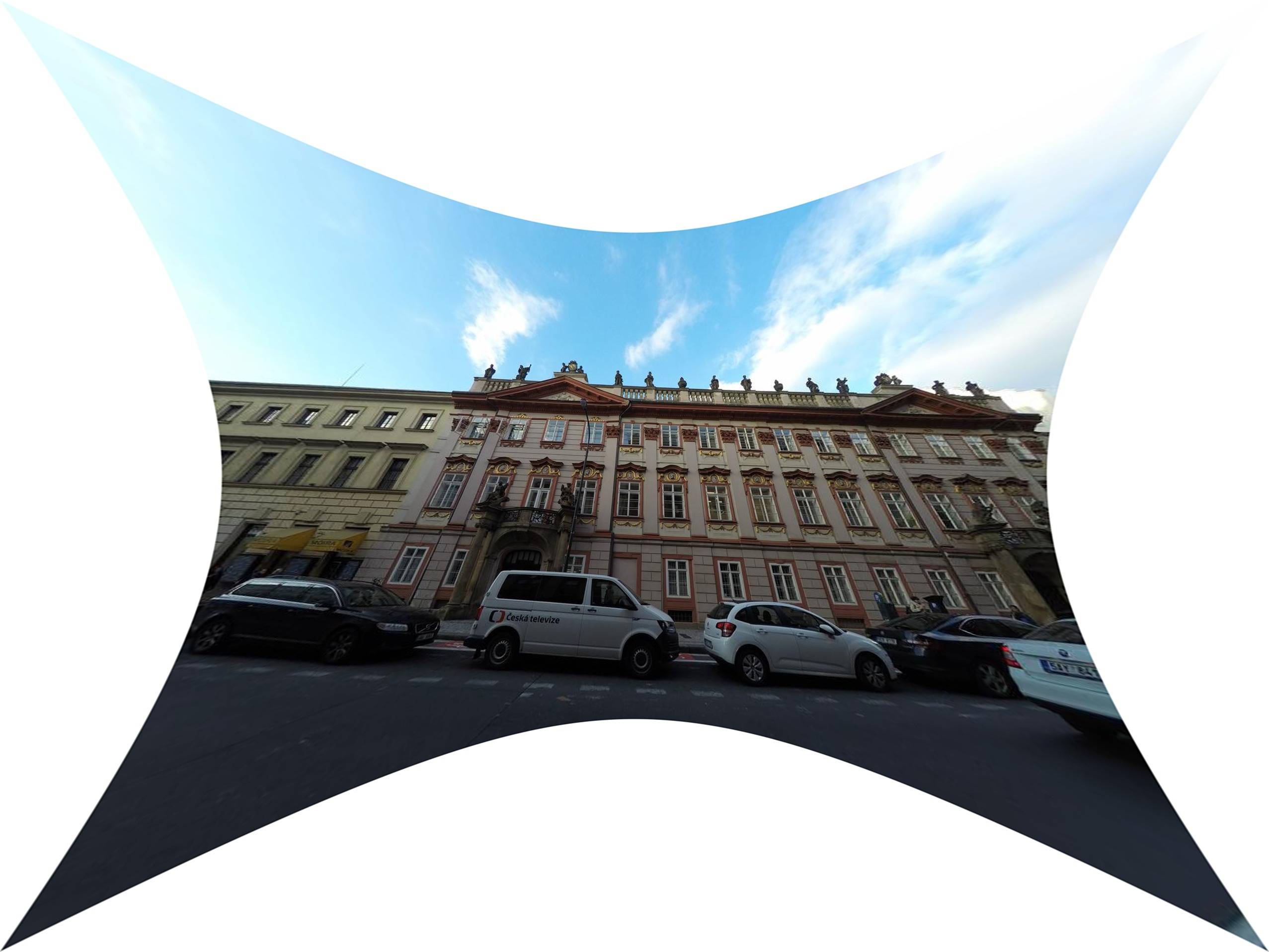}
\end{minipage}
\begin{minipage}{0.01\textwidth}
\hfill
\centering
\end{minipage}
\begin{minipage}{0.01\textwidth}
\hfill
\centering
\end{minipage}
\begin{minipage}{0.225\textwidth}
\centering
\includegraphics[width=\textwidth]{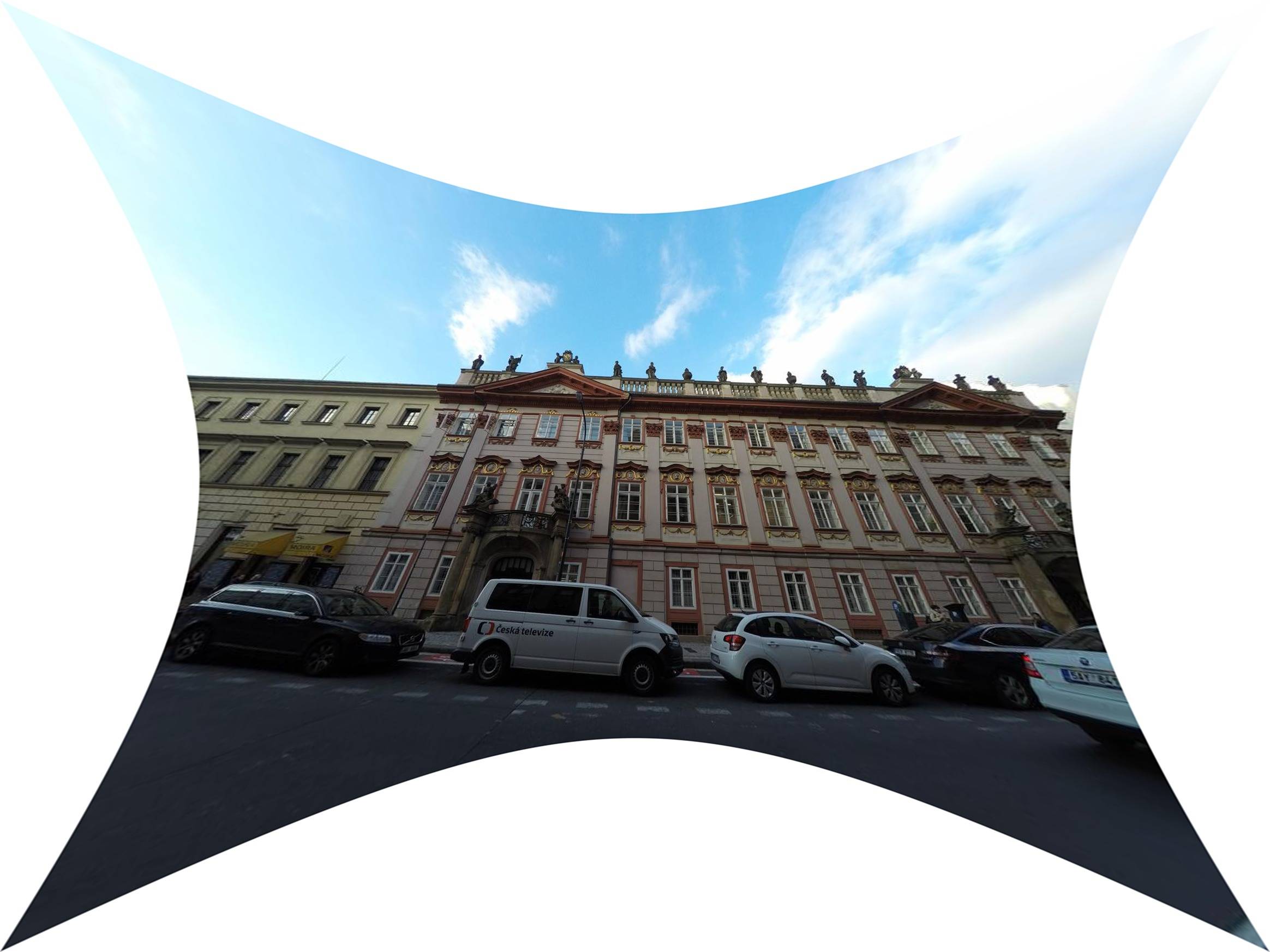}
\end{minipage}
\begin{minipage}{0.01\textwidth}
\hfill
\centering
\end{minipage}
\caption{\emph{GoPro Hero 4 at the wide setting for different solvers.} Results from \LORANSAC (see Sec.~\ref{sec:ransac}) for $\mH2\ve[l]\ve[u]$, which omits distortion, and the proposed solvers $\mH2.5\vl\vu\lambda$ and $\mH3.5\vl\vu\vv\lambda$. The top row has rectifications after local optimization (LO);  The bottom row has undistortions estimated from the best \emph{minimal} sample. \LORANSAC cannot recover from the poor initializations by $\mH2\ve[l]\ve[u]$ (column 2). The proposed solvers in columns 3 and 4 give a correct rectification. The bottom left has a chessboard undistorted using the division parameter estimated from the building facade by $\mH2.5\vl\vu\lambda$+LO.}
\label{fig:real_wide}
\end{figure*}

The proposed solvers---$\mH2.5\vl\vu\lambda,\mH3\vl\vu
s_{\vu}\lambda,\mH3.5\vl\vu\vv\lambda,$ $\mH4\vl\vu\vv
s_{\vv}\lambda$---estimate rectifications with less than 5 pixel RMS
warp error $\Delta^{\mathrm{warp}}_{\mathrm{RMS}}$, even at the 2
pixel noise level, see Fig.~\ref{fig:warp}. The need to model radial
lens distortion is shown by the biased fits for the solvers
$\mH2\vl\ve[u],\mH4\gamma$.

\section{Real Images}
\label{sec:real_images}
In the qualitative experiments on real images shown in
Figs.~\ref{fig:first} and ~\ref{fig:gopro}, we tested the proposed
solvers on GoPro4 Hero 4 images with increasing field-of-view
settings, namely narrow, medium and wide, where a wider field-of-view
setting generates more extreme radial distortion since the boundary of
the lens is used. The proposed method generates high-quality
rectifications at all the field-of-view settings. More real-image
experiments, including results for cameras with radial distortions
that are typical for mobile phone cameras and fisheye lenses (\eg, 8mm
lens) can be found in Sec.~\ref{sec:extended_experiments} in the
supplementary material.

The experiment shown in Fig.~\ref{fig:real_wide} compares the performance
of two of the proposed solvers, $\mH2.5\vl\vu\lambda$ and
$\mH3.5\vl\vu\vv\lambda$, to $\mH2\ve[l]\ve[u]$ in a state-of-the-art
local-optimization (LO) framework (see Sec.~\ref{sec:ransac}) on an
GoPro Hero 4 image at the wide field-of-view setting. The two proposed
solvers accurately estimate the division-model parameter (see the
undistorted reference chessboard in Fig.~\ref{fig:real_wide}) and the
rectification, while the LO-variant using the $\mH2\ve[l]\ve[u]$
solver is unable to recover the lens distortion parameter. See
Fig.~\ref{fig:real} in the supplemental material for results on an
image at the medium field-of-view setting.

\section{Conclusions}
This paper proposes the first minimal solvers that jointly solve for
the affine rectification of an imaged scene plane and a camera's
radial lens distortion from coplanar repeated patterns.  Rectification
and radial lens distortion are recovered from only one conjugately
translated affine-covariant feature or two independently translated
similarity-covariant features. Synthetic experiments demonstrate the
good stability and superior robustness to noise with respect to
measures of rectification accuracy and lens-distortion estimation of
the proposed solvers as compared to the state-of-the-art. However, the
polynomial constraint equations that arise from conjugate translations
distorted by the division model need to be transformed with the
hidden-variable trick to generate stable solvers, though. Qualitative
real-image experiments demonstrate high-quality rectifications for
highly-distorted wide-angle lenses, which was not possible using the
state-of-the-art. Future work could include conditionally sampling the
measurements during robust estimation to take into account their size,
relative distance from each other, or distance from the distortion
center. We expect these factors have a big impact on rectification
quality, but this study was beyond scope for this paper.

%
%

\section{Acknowledgements}
James Pritts was supported by the grants MSMT LL1303 ERC-CZ and
SGS17/185/OHK3/3T/13, Zuzana Kukelova by the Czech Science Foundation
Project GACR P103/12/G084, Viktor Larsson by the strategic research
projects ELLIIT and eSSENCE, Swedish Foundation for Strategic Research
project ”Semantic Mapping and Visual Navigation for Smart Robots”
(grant no. RIT15-0038) and Wallenberg Autonomous Systems and Software
Program (WASP), and Ondrej Chum by the grant MSMT LL1303 ERC-CZ.

{\small
  \bibliographystyle{ieee}

}

\appendix
\numberwithin{figure}{section}
\newpage
\addtocontents{toc}{\protect\setcounter{tocdepth}{2}}
\pagestyle{plain}

\setcounter{figure}{0}
\setcounter{table}{0}
\counterwithin{figure}{section}
\counterwithin{table}{section}


\section{Transfer Error} 
\label{sec:xfer_error_details}
The scene plane is tessellated by a 10x10 square grid of points,
denoted $\{\,\ve[X]_i\,\}$, with a 1 meter spacing between adjacent
points. Suppose that $\ve[y] \leftrightarrow \ve[y]^{\prime}$ is an
undistorted point correspondence induced by the conjugate translation
$\mH_{\vu}=[\ma{I}_3+\vu\vl^{\T}]$ in the imaged scene plane (here we
assume that $s^{\vu}_i=1$ since we speak about an individual point
correspondence).

Points $\{\,\ve[X]_i\,\}$ are translated by 1 meter on the scene plane
in the direction of translation induced by the preimage of the point
correspondence $\ve[y] \leftrightarrow \ve[y]^{\prime}$ giving the
translated grid $\{\,\ve[X]^{\prime}_i\,\}$. The purpose of
constructing the grid and it's translation is to uniformly cover the
scene plane that the camera images in its field of view. In this way,
the accuracy of the conjugate translation and lens-distortion
parameter estimation can be measured across most of the image. The
conjugate translation $\mH_{\vu}$ is not used directly because the
magnitude of translation may span the extent of the scene plane, so
applying it to the tessellation would transform the grid out of the
field of view.

Let the camera be parameterized by the camera matrix
$\mP=(\mA\mH)^{\inv}$ (see Sec.~\ref{sec:warp_error} for the
definition of the camera matrix) that pointwise maps the scene plane
$\Pi$ to the imaged scene plane $\pi$ and division model parameter
$\lambda$. The preimages of the undistorted point correspondence
$\ve[y]
\leftrightarrow \ve[y]^{\prime}$ in the scene-plane coordinate system
is, respectively, $\beta \ve[Y]=\mP^{\inv}\ve[y]$ and
$\beta^{\prime} \ve[Y]^{\prime}= \mP^{\inv}\ve[y]^{\prime}$. The
translation \ve[t] of the preimages in the scene plane coordinate
system is
$\ve[t]=\ve[Y]-\ve[Y]^{\prime}=\rowvec{3}{t_x}{t_y}{0}^{\T}$.

Then $\|\ve[t]\|$ is the magnitude of translation between the repeated
scene elements in the scene-plane coordinate system. Denote the
homogeneous translation matrix $\ma{T}(\ve[t])$ to be the matrix
constructed from \ve[t] as
\begin{equation}
\ma{T}(\ve[t]) = \begin{bmatrix}
1 & 0 & t_x \\ 0 & 1 & t_y \\ 0 & 0 & 1
\end{bmatrix}.
\end{equation}
The translation of the grid points by unit distance is given by
$\vX^{\prime}=\ma{T}(\ve[t]/\|\ve[t]\|)\vX$. Recall
from \eqref{eq:conjugate_translation} that a conjugate translation has
the form $\mP\ma{T(\cdot)}\mP^{\inv}$. Using \eqref{eq:decomposition},
the conjugate translation of unit distance in the direction of point
correspondences $\ve[y] \leftrightarrow \ve[y]^{\prime}$ is
\begin{align*}
\mH_{\ve[u]/\|\ve[t]\|} &= \mP\ma{I}_3\mP^{\inv}+\mP\colvec{3}{t_x/\|\ve[t]\|}{t_y/\|\ve[t]\|}{1}\left[\mP^{-\T}\colvec{3}{0}{0}{1}\right]^{\T} \\
&=
[\ma{I}_3+\frac{\ve[u]}{\|\ve[t]\|}\ve[l]^{\T}] \numberthis \label{eq:scaled_transfer_error}.
\end{align*}

The unit conjugate translation $\mH_{\ve[u]/\|\ve[t]\|}$ can be
written in terms of the conjugate translation $\mH_{\vu}$ induced by
the undistorted point correspondence
$\ve[y] \leftrightarrow \ve[y]^{\prime}$ as
\begin{align*}
\ma{I}_3+\frac{\ve[u]}{\|\ve[t]\|}\ve[l]^{\T} &= \ma{I}_3+\frac{1}{\|\ve[t]\|}[\ma{I}_3+\ve[u]\ve[l]^{\T}-\ma{I}_3] \\
&= \ma{I}_3+\frac{1}{\|\ve[t]\|}[\mH_{\ve[u]}-\ma{I}_3] \numberthis \label{eq:transfer_error_derived}.
\end{align*}
The derivation of \eqref{eq:transfer_error_derived} gives the form of
transformation used in the transfer error
$\Delta_{\mathrm{RMS}}^{\mathrm{xfer}}$ defined in
Sec.~\ref{sec:transfer_error}, which maps from the undistorted points
of the grid $\{ \vx \}$ to their translated correspondences
$\{ \vx^{\prime} \}$.

\section{Computational Complexity}  
\label{sec:wall_clock}
Table~\ref{table:tpl_size} lists the elimiation template sizes for the
proposed solvers. The average time to compute the solutions for a
given input for a solver is directly proportional to the elimination
template size. The solvers are implemented in MATLAB and
C++. Significantly faster implementations are possible with a pure C++
port. Still the solvers are sufficiently fast for use in \RANSAC. The
proposed solvers have an average solve time from 0.3 to 2
milliseconds.
\begin{table}[H] \centering
\ra{1}
  \begin{tabular}{@{} c c c c @{}}
\toprule              
   $\mH2.5\vl\vu\lambda$ & $\mH3\vl\vu s_{\vu}\lambda$ & $\mH3.5\vl\vu\vv\lambda$ & $\mH4\vl\vu\vv s_{\vv}\lambda$ \\
\midrule 
\small{14x18} & \small{24x26} & \small{54x60} & \small{76x80} \\
\bottomrule
\end{tabular} 
\caption{Template sizes for the proposed solvers.}
\label{table:tpl_size} 
\end{table}


%

\section{Extended Experiments} 
\label{sec:extended_experiments} 
The extended real-data experiments in the following pages include
\begin{enumerate*}[(i)]\item images with lesser radial distortion from consumer cameras and mobile phones; the images also demonstrate the proposed method's effectiveness on diverse
scene content, \item images for very wide field-of-view lenses (8mm
and 12mm), \item and an additional local-optimization experiment
similar to the one in Fig.~\ref{fig:real_wide} on a GoPro Hero 4 image
taken with its medium field-of-view setting, which further
demonstrates the need for a minimal solver that jointly estimates lens
distortion with affine-rectification to achieve an accurate
undistortions and rectifications.
\end{enumerate*}

\newpage

\newcommand{\todo}[1]{\textbf{\color{red}#1}}
\newcommand{\bmat}[1]{\begin{bmatrix}#1\end{bmatrix}}
\newcommand{\pmat}[1]{\begin{pmatrix}#1\end{pmatrix}}
\renewcommand{\vec}[1]{\textbf{#1}}

\begin{figure*}
\newcommand{\trirow}[1]{%
\setlength{\tabcolsep}{-0.06cm}
\begin{tabular}{c}
\includegraphics[height=2.5cm]{suppimg/#1.jpg}\\
\includegraphics[height=2.5cm]{suppimg/#1_ud.jpg}\\
\includegraphics[height=2.3cm]{suppimg/#1_rect_cropped.jpg}\\
\end{tabular}
}

\trirow{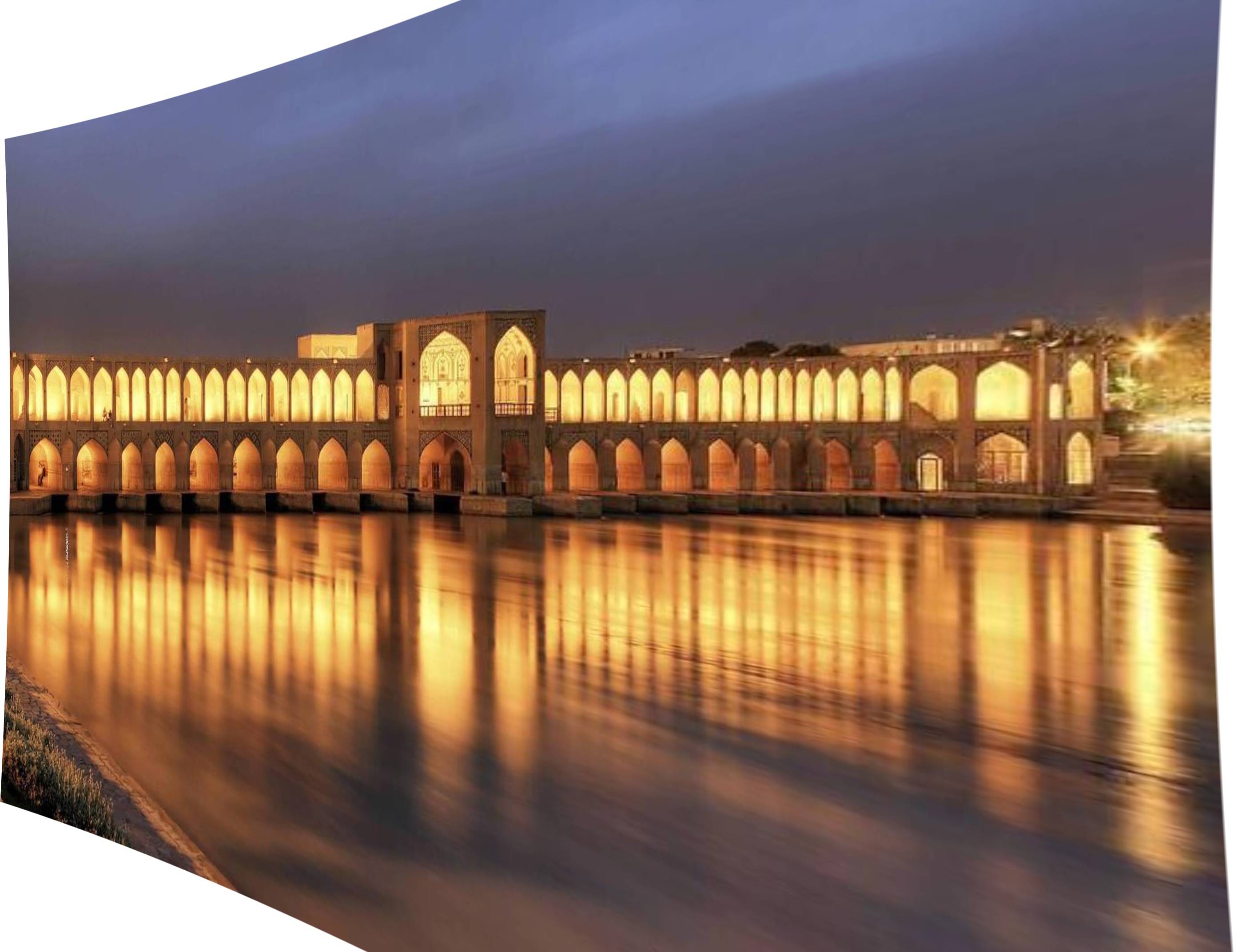}
\trirow{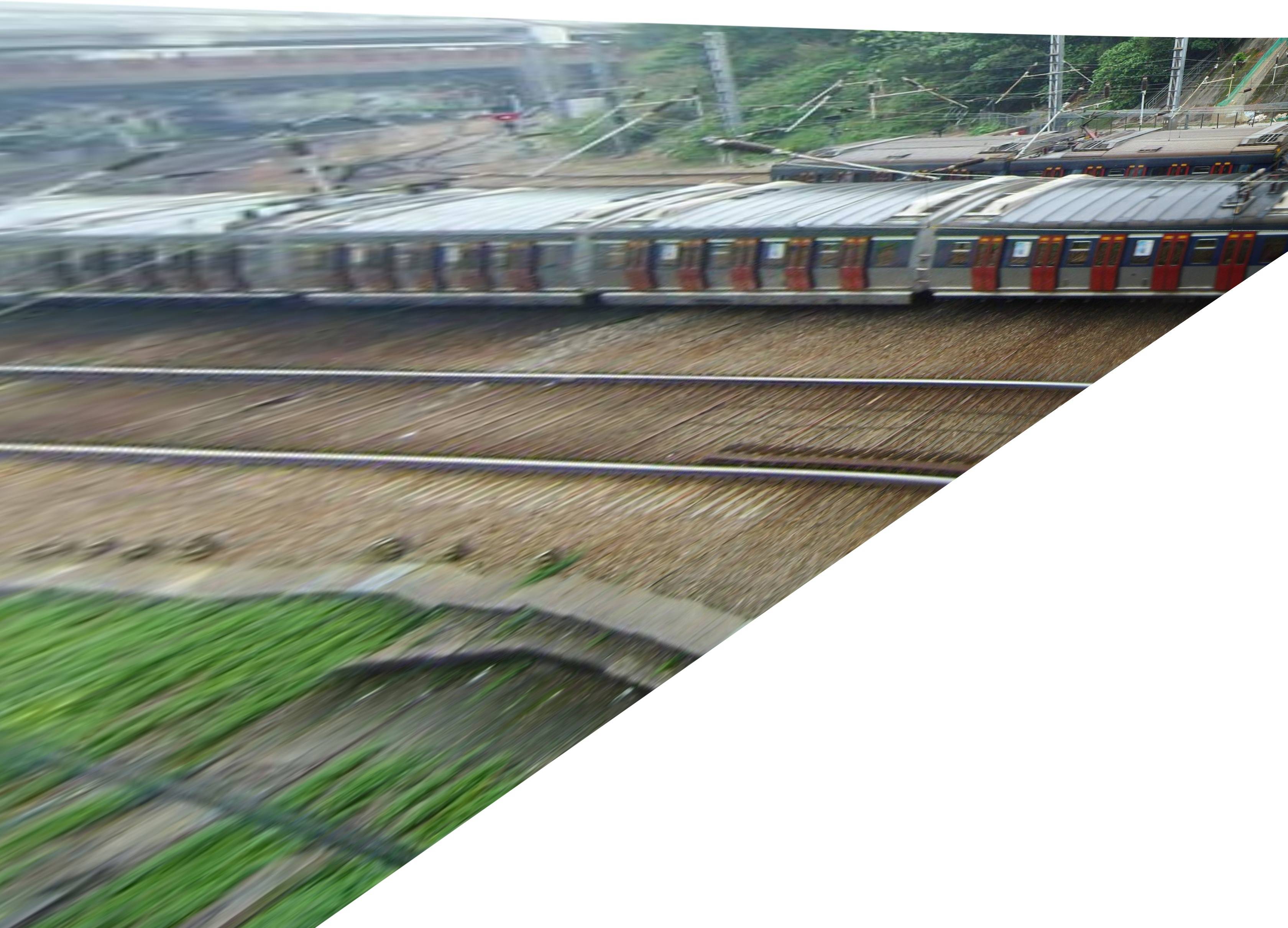}
\trirow{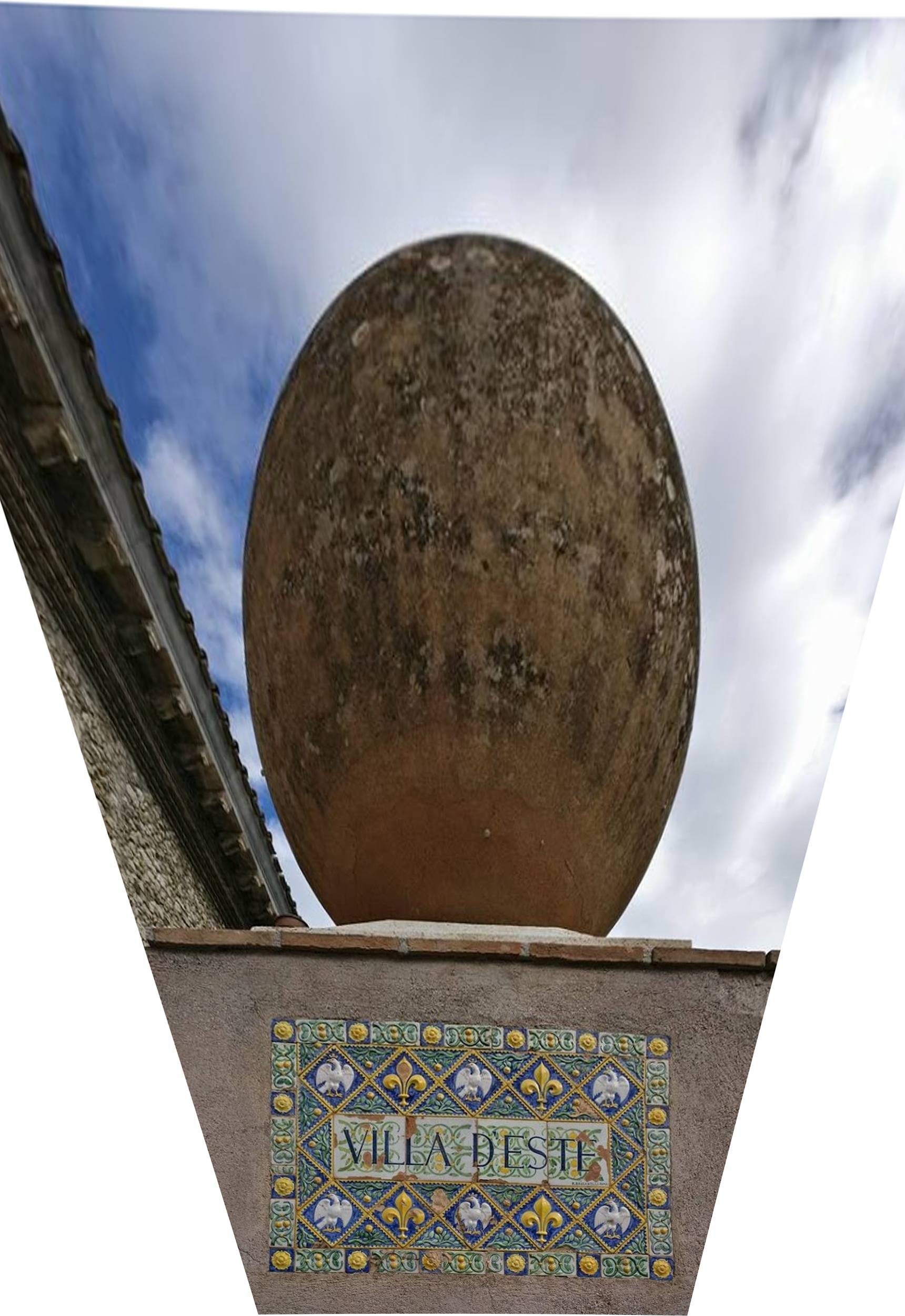}
\trirow{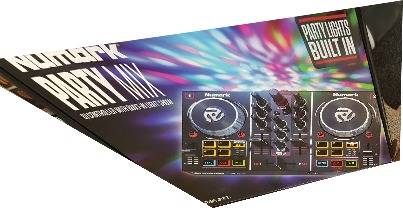}
\trirow{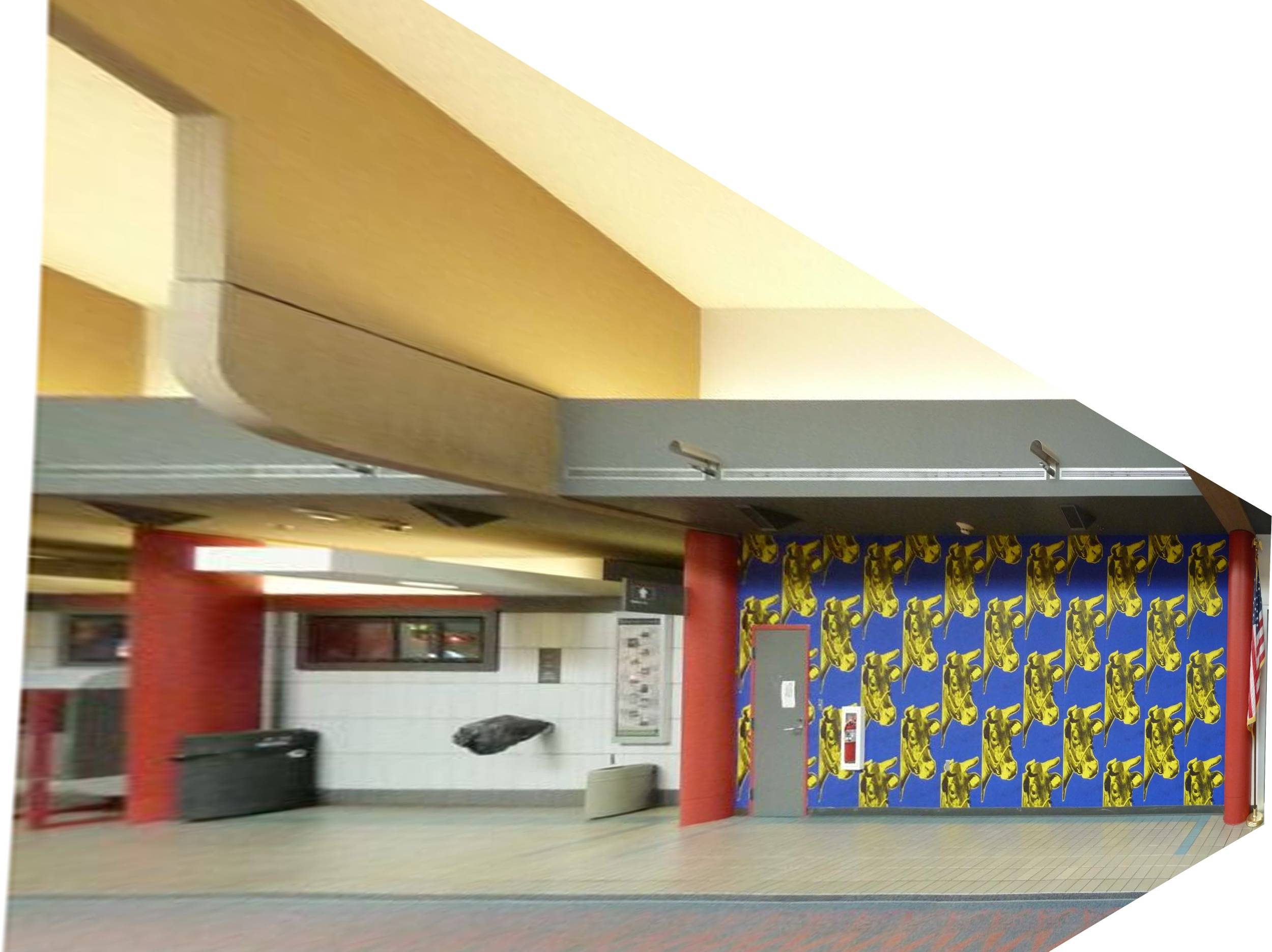}
\caption{\emph{Narrow field-of-view and diverse scene-content
    experiments for} $\mH2.5\vl\vu\lambda$+LO. The proposed method
  works well if the input image has little or no radial lens
  distortion. This imagery is typical of consumer cameras and mobile
  phone cameras. The images are diverse and contain unconventional
  scene content. Input images are on the top row; undistorted images
  are on the middles row, and the rectified images are on the bottom
  row.}
\end{figure*}

\begin{figure*}[t!]
\begin{minipage}{0.22\textwidth}
\centering
Original image
\includegraphics[width=\textwidth]{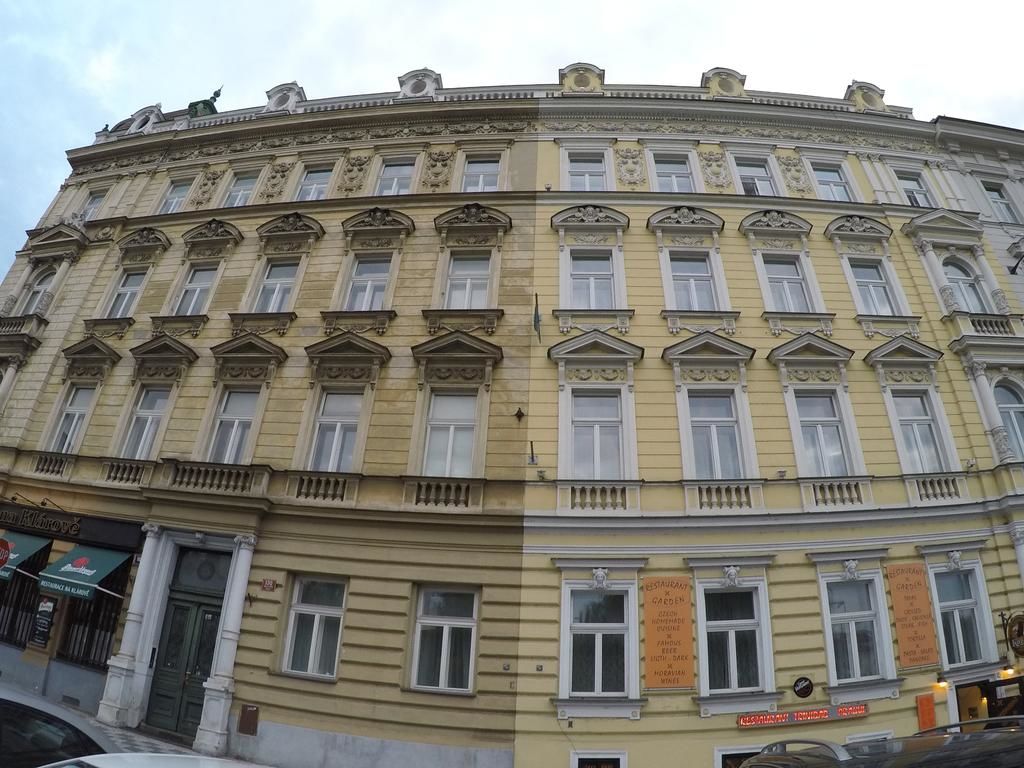}
\end{minipage}
\begin{minipage}{0.245\textwidth}
\centering
$\mH2\ve[l]\ve[u]$+LO; $20.3 \%$ inliers
\includegraphics[width=\textwidth]{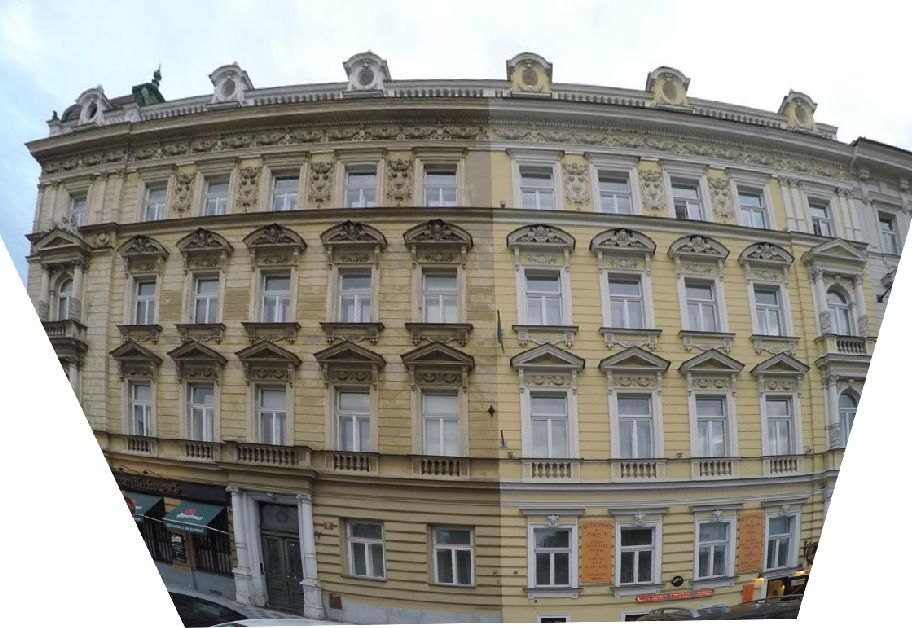}
\end{minipage}
\begin{minipage}{0.245\textwidth}
\centering
$\mH2.5\vl\vu\lambda$+LO; $31.2 \%$ inliers
\includegraphics[width=\textwidth]{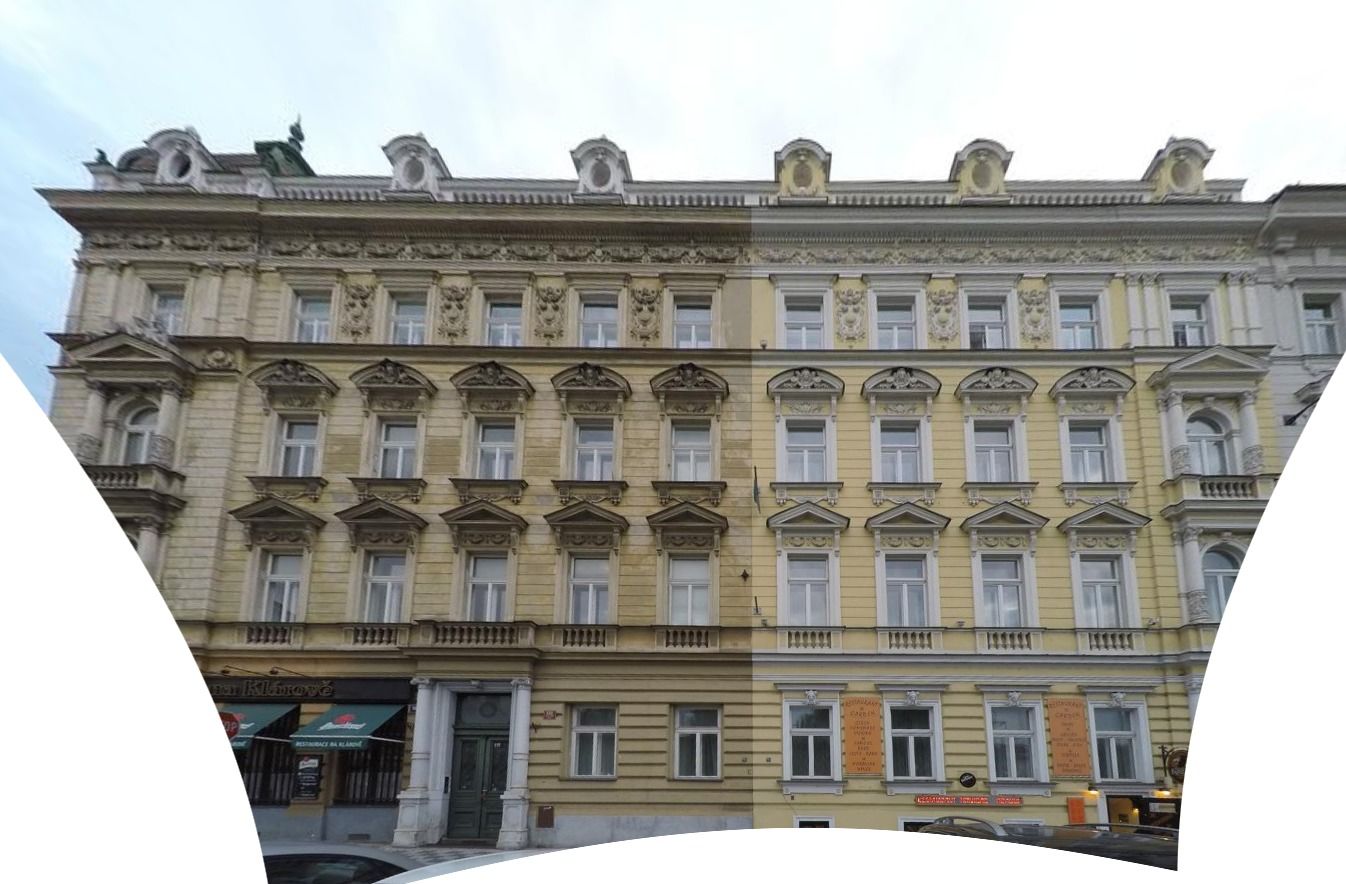}
\end{minipage}
\begin{minipage}{0.245\textwidth}
\centering
$\mH3.5\vl\vu\vv\lambda$+LO; $30.7 \%$ inliers
\includegraphics[width=\textwidth]{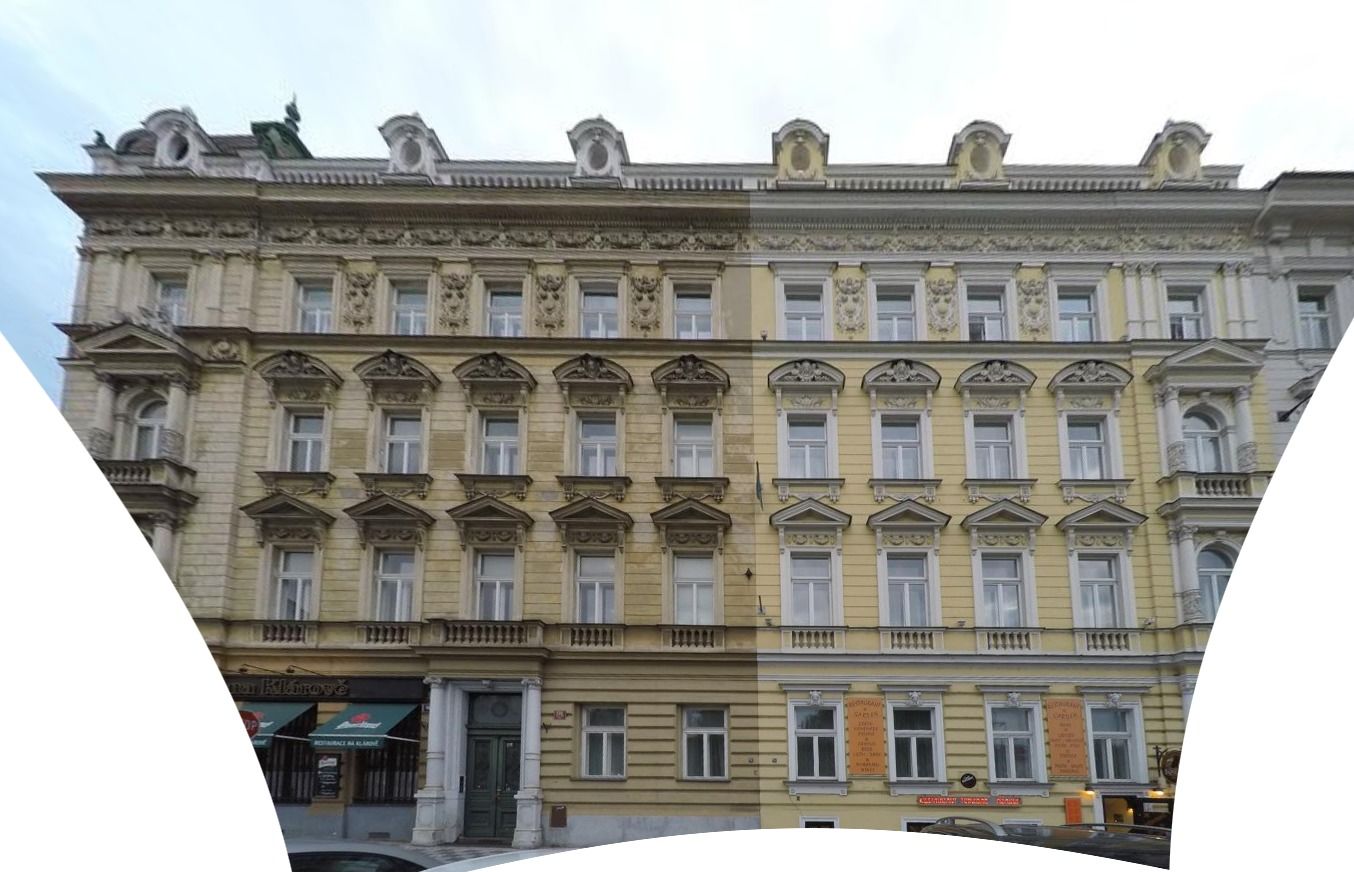}
\end{minipage}

\begin{minipage}{0.22\textwidth}
\centering
\includegraphics[width=\textwidth]{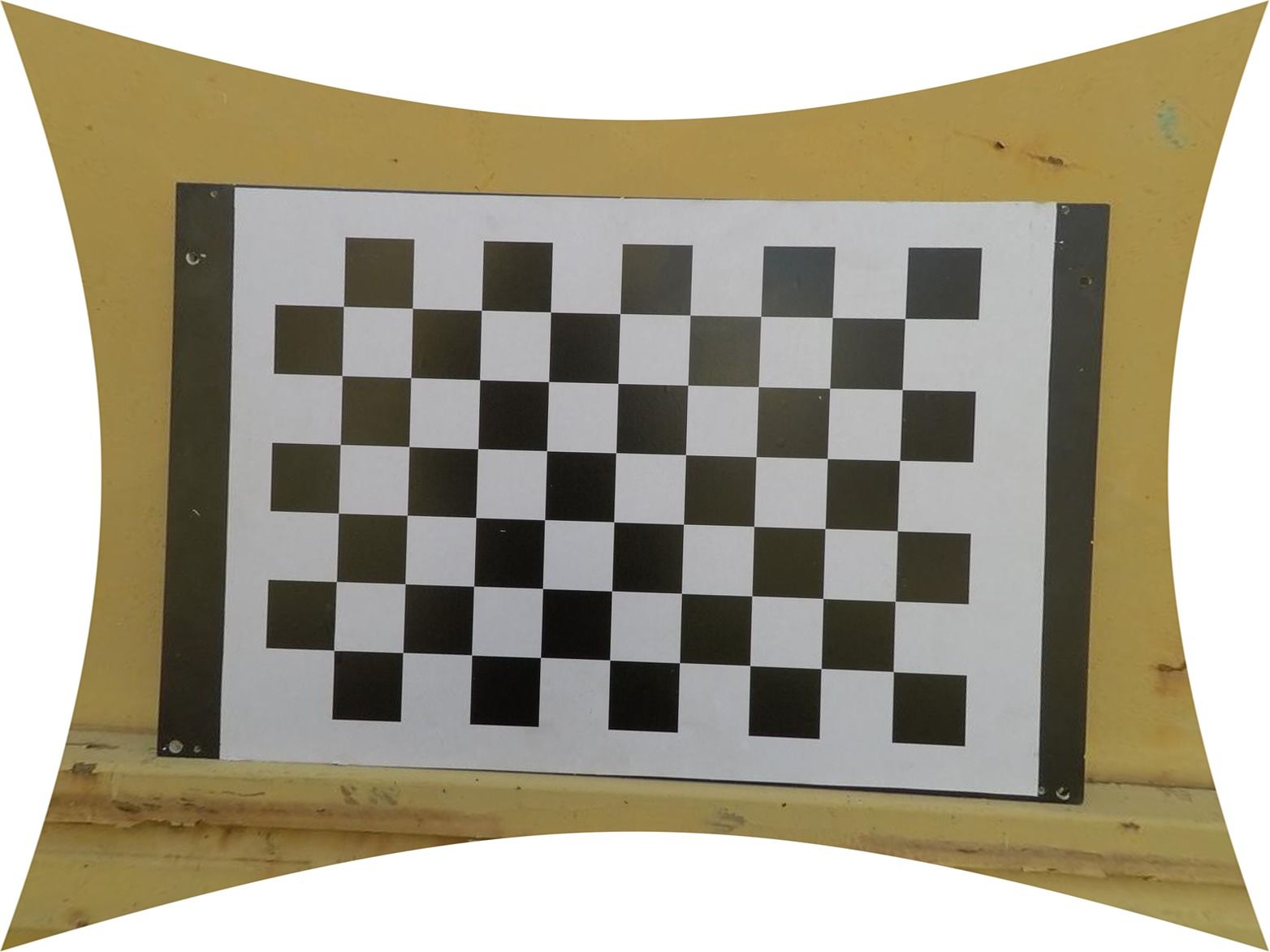}
\end{minipage}
\begin{minipage}{0.01\textwidth}
\hfill
\centering
\end{minipage}
\begin{minipage}{0.225\textwidth}
\centering
\includegraphics[width=\textwidth]{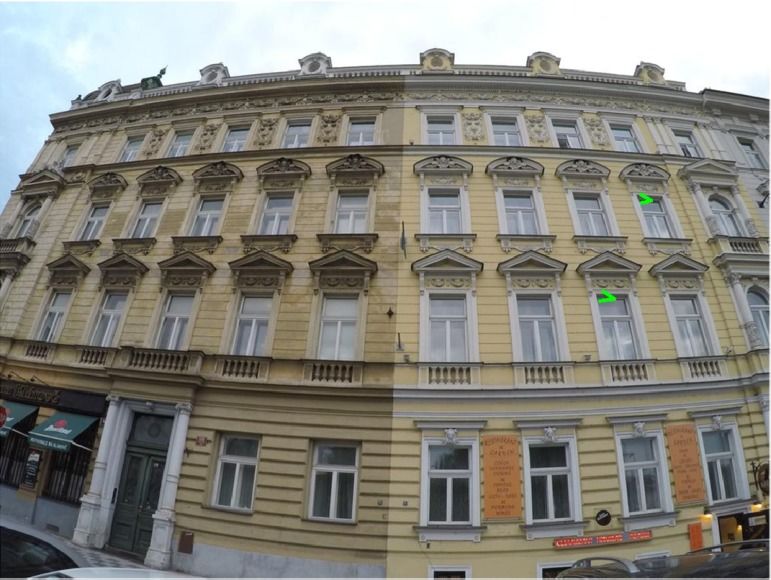}
\end{minipage}
\begin{minipage}{0.01\textwidth}
\hfill
\centering
\end{minipage}
\begin{minipage}{0.01\textwidth}
\hfill
\centering
\end{minipage}
\begin{minipage}{0.225\textwidth}
\centering
\includegraphics[width=\textwidth]{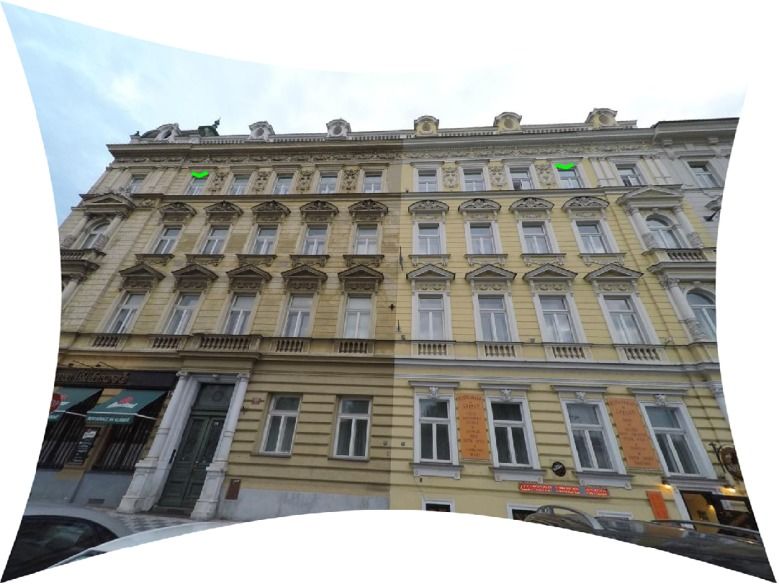}
\end{minipage}
\begin{minipage}{0.01\textwidth}
\hfill
\centering
\end{minipage}
\begin{minipage}{0.01\textwidth}
\hfill
\centering
\end{minipage}
\begin{minipage}{0.225\textwidth}
\centering
\includegraphics[width=\textwidth]{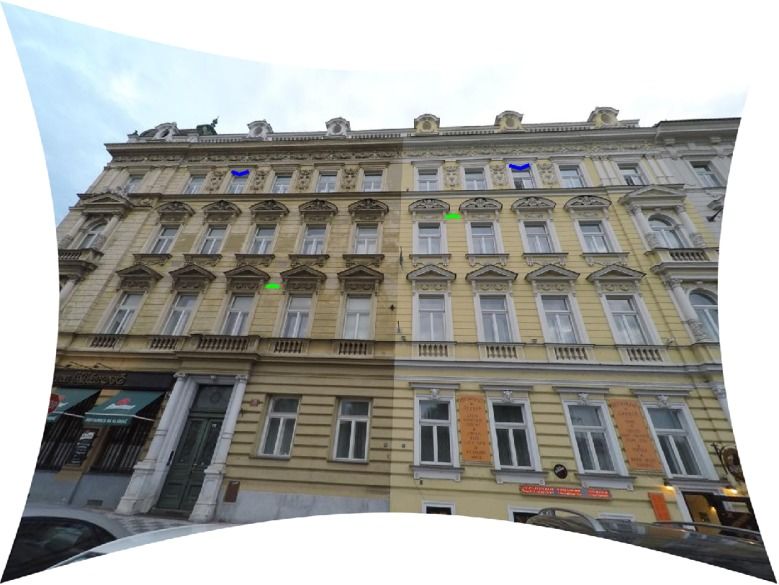}
\end{minipage}
\begin{minipage}{0.01\textwidth}
\hfill
\centering
\end{minipage}
\caption{\emph{GoPro Hero 4 at the medium setting for different
    solvers.} Results from \LORANSAC (see Sec.~\ref{sec:ransac}) for
    $\mH2\ve[l]\ve[u]$, which omits distortion, and the proposed
    solvers $\mH2.5\vl\vu\lambda$ and $\mH3.5\vl\vu\vv\lambda$. The
    top row has rectifications after local optimization (LO); The
    bottom row has undistortions estimated from the best
    \emph{minimal} sample. \LORANSAC fails from the poor
    initializations by $\mH2\ve[l]\ve[u]$ (column 2). The proposed
    solvers in columns 3 and 4 give a correct rectification. The
    bottom left has a chessboard undistorted using the division
    parameter estimated from the building facade by
    $\mH2.5\vl\vu\lambda$+LO.}
\label{fig:real}
\end{figure*}

\begin{figure*}
\newcommand{\trirow}[1]{%
\setlength{\tabcolsep}{0.4cm}
\begin{tabular}{c}
\includegraphics[height=5.5cm]{suppimg/#1.jpg}\\
\includegraphics[height=5.5cm]{suppimg/#1_ud.jpg}\\
\includegraphics[height=5.5cm]{suppimg/#1_rect_cropped.jpg}\\
\end{tabular}
}

\trirow{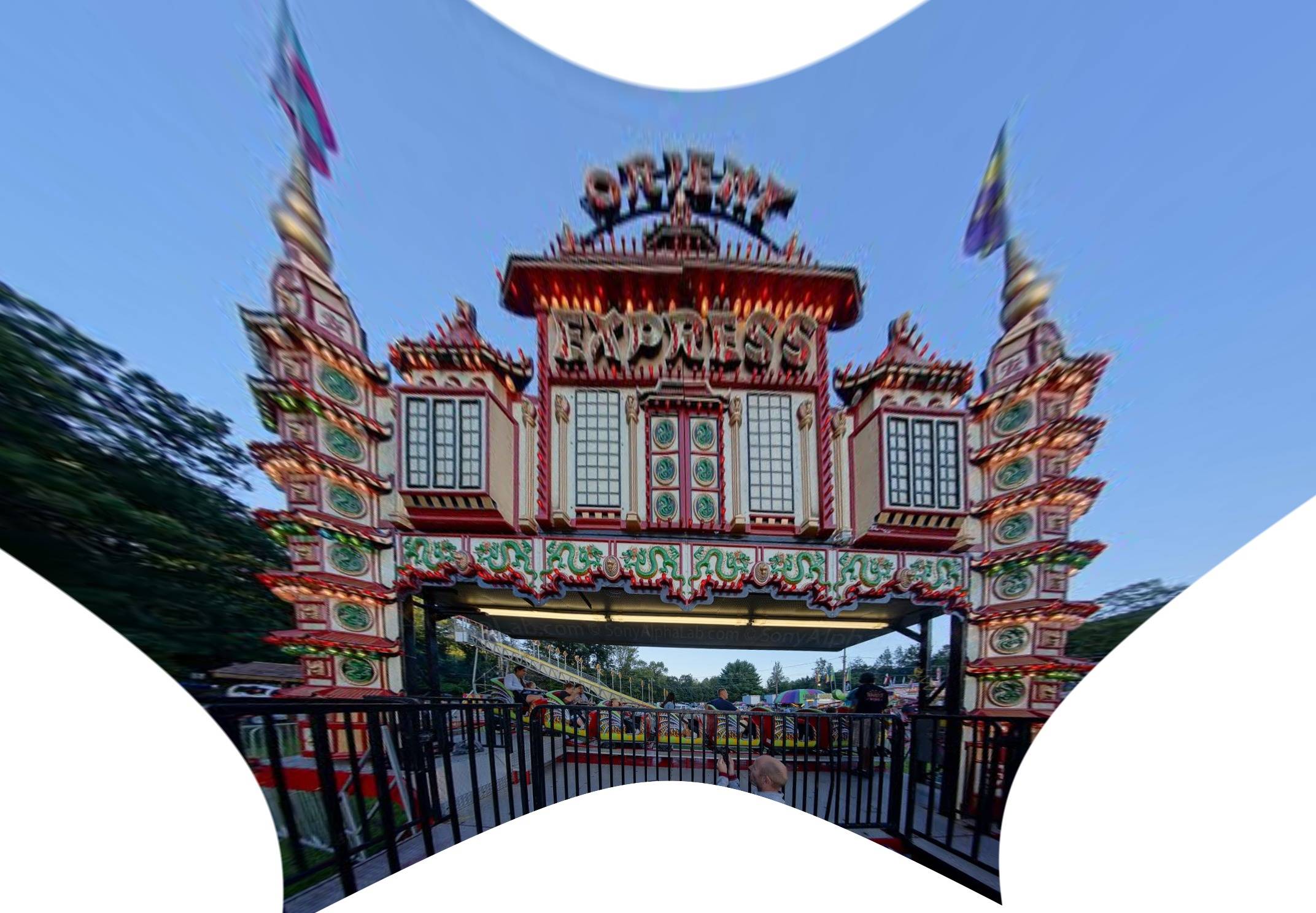}
\trirow{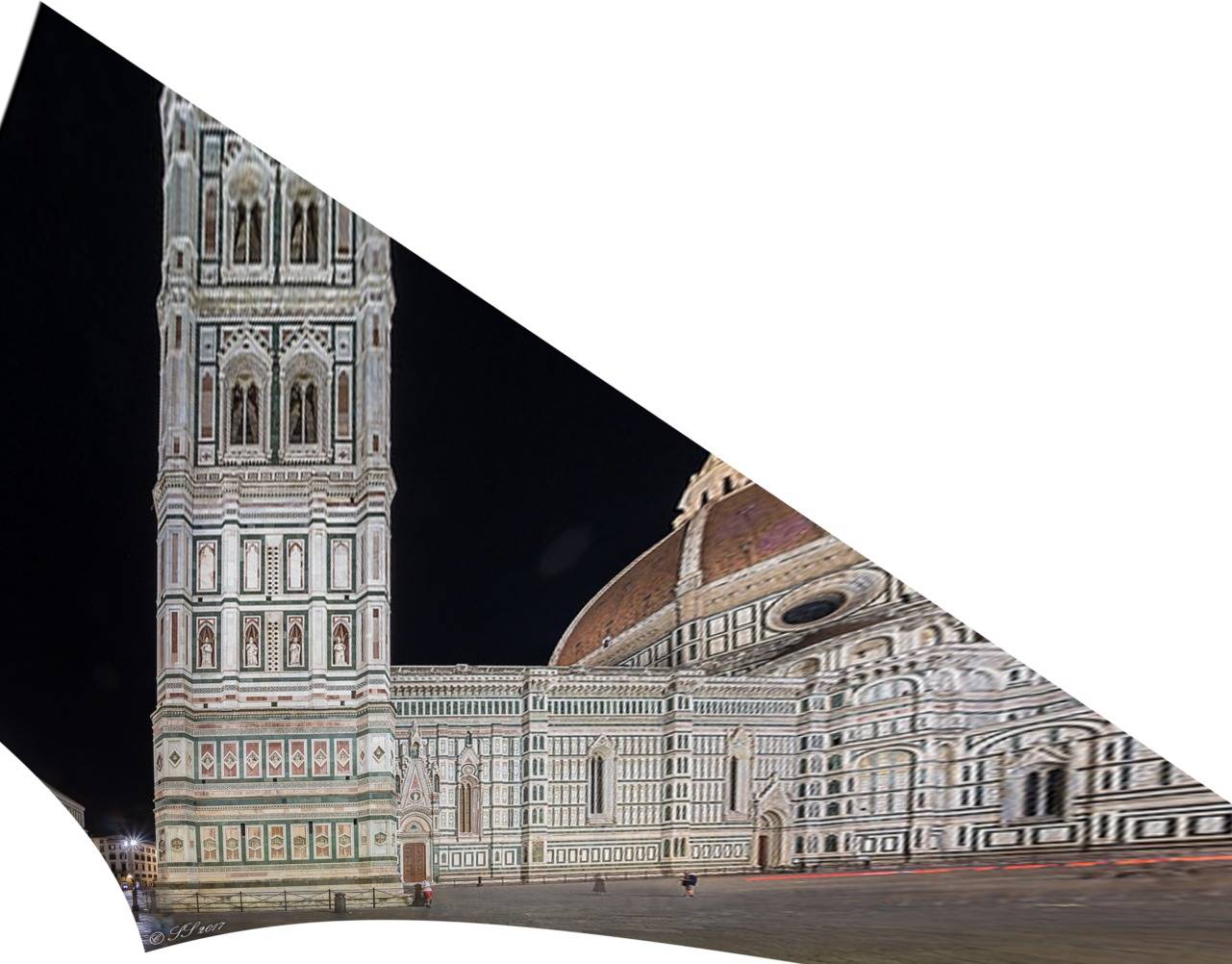}
\caption{\emph{Very wide-angle images undistorted and rectified with}
  $\mH2.5\vl\vu\lambda$+LO. The left column is an image from an 8mm
  lens, and the right column is from a 12mm lens. The top row contains
  the input images; the middle row contains the undistorted images,
  and the bottom row contains the rectified images. The division model
  \cite{Fitzgibbon-CVPR01} used for radial lens distortion has only 1
  parameter, which may impose limits for modeling extreme lens
  distortion.}
\end{figure*}

\begin{figure*}
\centering
\begin{minipage}{0.495\textwidth}
\centering
Affine-covariant feature detections
\includegraphics[width=\linewidth]{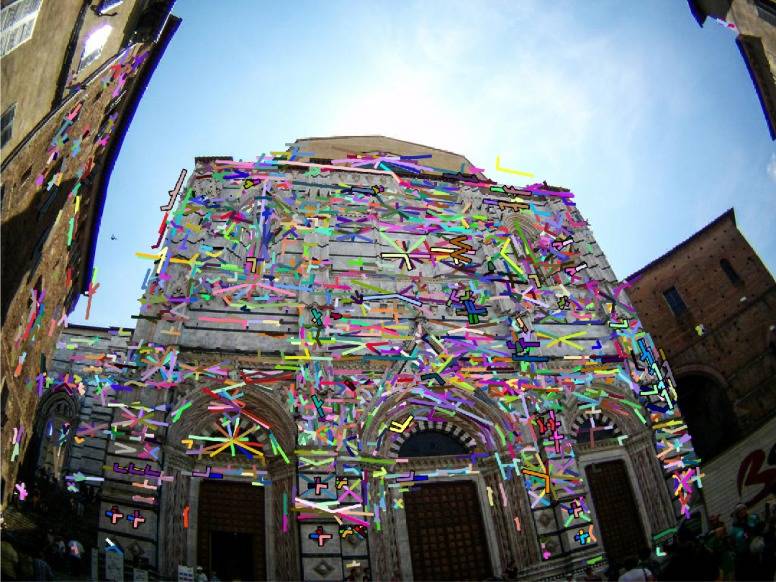}
\end{minipage}
\begin{minipage}{0.495\textwidth}
\centering
Inlying affine-covariant features
\includegraphics[width=\linewidth]{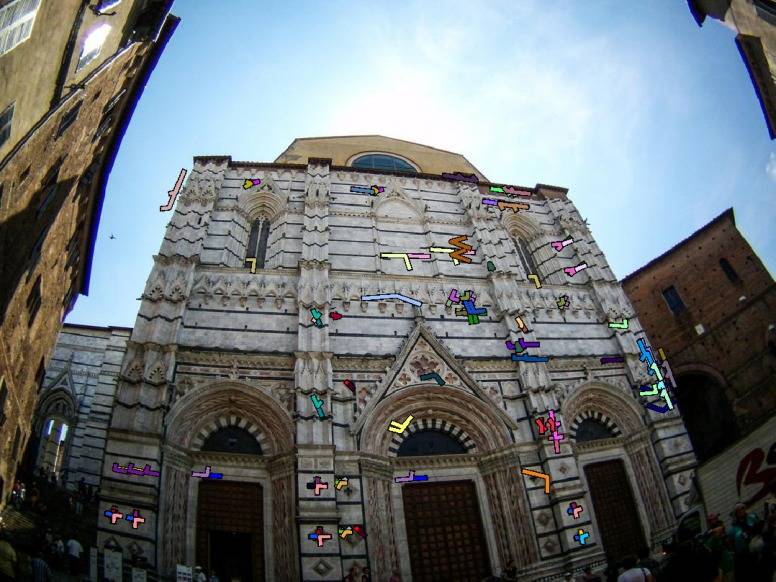}
\end{minipage}
\begin{minipage}{0.495\textwidth}
\centering
Undistorted
\includegraphics[width=\textwidth]{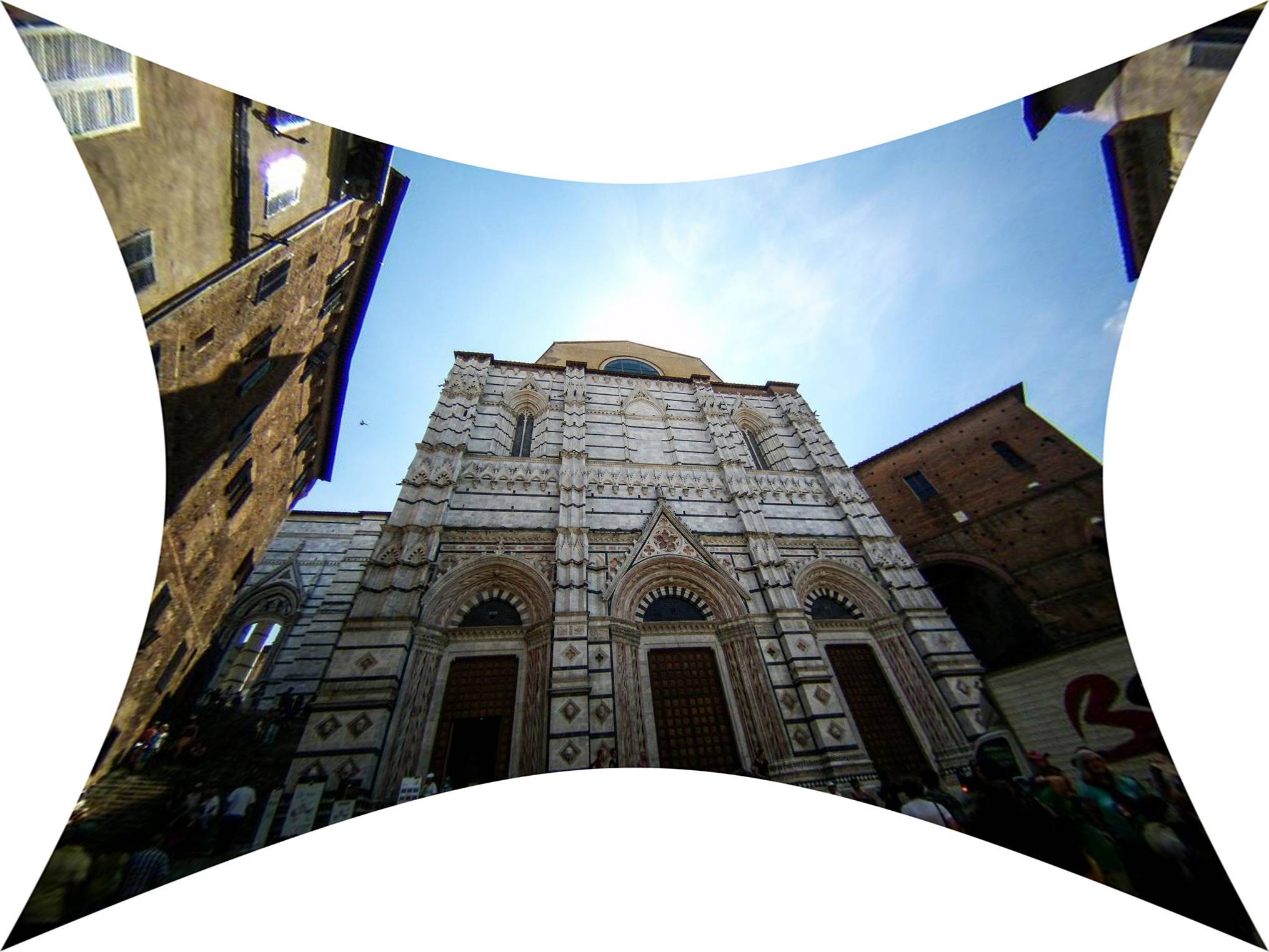}
\end{minipage}
\begin{minipage}{0.495\textwidth}
\centering
Undistorted and rectified
\includegraphics[width=\textwidth]{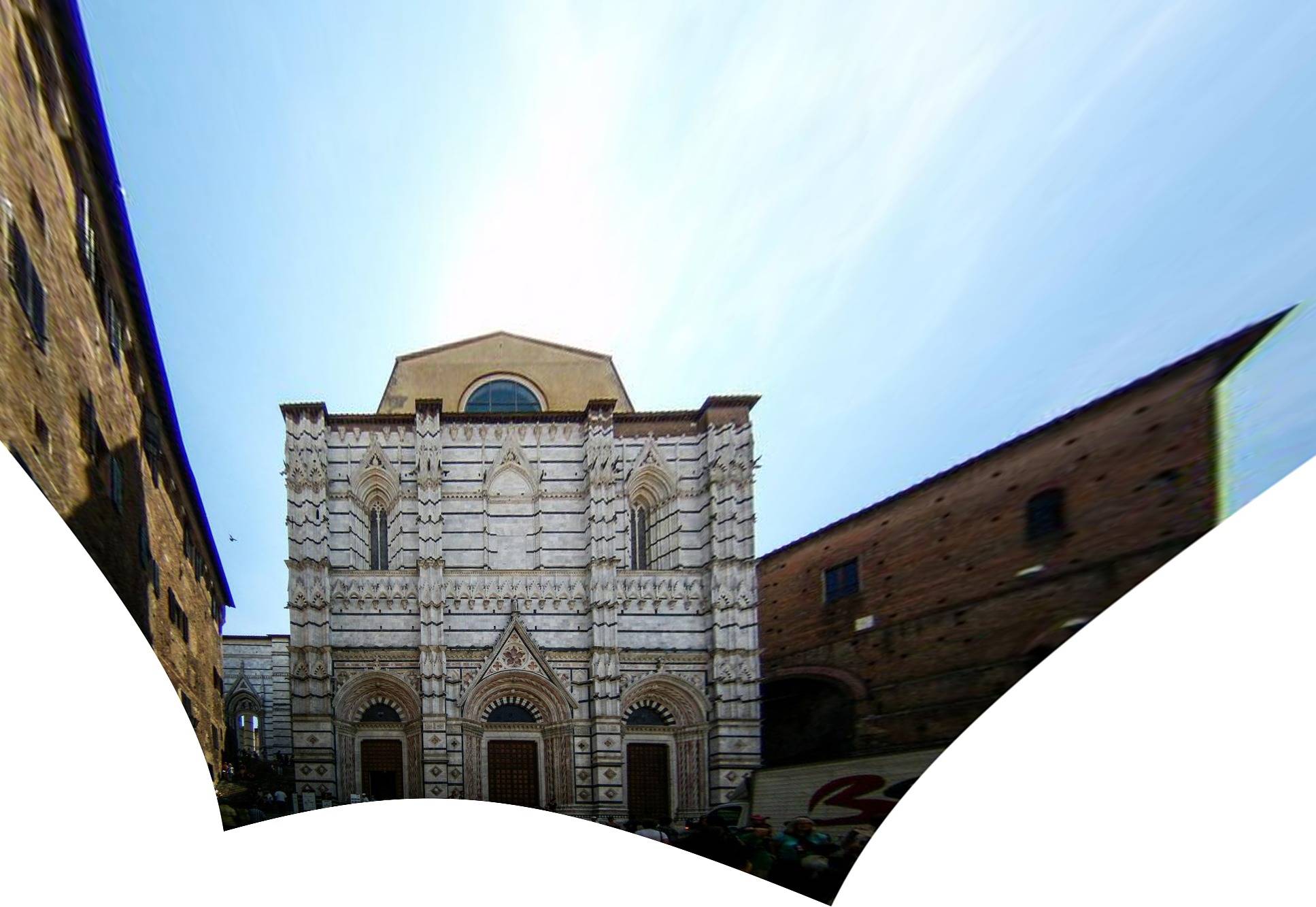}
\end{minipage}
\caption{\emph{Problem difficulty and method robustness} The input to
  the method is ungrouped affine-covriant features (top left). Common
  problems include missed detections of repeated texture, duplicate
  detections, and detections due to compression artifacts, all of
  which are visible in this example. The inliers with respect to the
  total number of detected features can be a very small proportion
  (top right). Still the method can estimate accurate undistortion
  (bottom left) and affine rectification (bottom right) even from a
  very sparse sampling of the inlying affine-covariant features on the
  scene plane (top right).}
\label{fig:problem_difficulty}
\end{figure*}

\end{document}